\documentclass[journal]{IEEEtran}

\usepackage{graphics} 
\usepackage{epsfig} 
\usepackage{amsmath} 
\usepackage{amssymb}  
\usepackage{soul}
\usepackage{multirow}
\usepackage[bookmarks=false]{hyperref}

\begin{document}

\title{Simultaneous Traffic Sign Detection\\and Boundary Estimation\\using Convolutional Neural Network}

\author{Hee Seok Lee and Kang Kim
\thanks{\copyright IEEE. Personal use of this material is permitted. Permission from IEEE must be obtained for all other users, including reprinting/republishing this material for advertising or promotional purposes, creating new collective works for resale or redistribution to servers or lists, or reuse of any copyrighted components of this work in other works. DOI: 10.1109/TITS.2018.2801560. \textit{(Corresponding Author: Hee Seok Lee)}

Hee Seok Lee and Kang Kim are with Qualcomm Technologies Inc., Seoul, 06060, Korea {\tt\small (\{heeseokl, kangk\}@qti.qualcomm.com)}.}%
}

\markboth{\copyright IEEE. ACCEPTED FOR PUBLICATION IN IEEE TRANSACTIONS ON INTELLIGENT TRANSPORTATION SYSTEMS}%
{Shell \MakeLowercase{\textit{et al.}}: Bare Demo of IEEEtran.cls for IEEE Journals}

\maketitle

\def \ie {\textit{i.e.,}\quad}
\def \eg {\textit{e.g.,}\quad}
\def \bfp {\mathbf{p}}
\def \bfb {\mathbf{b}}
\def \bfc {\mathbf{c}}
\def \bfq {\mathbf{q}}

\begin{abstract}
We propose a novel traffic sign detection system that simultaneously estimates the location and precise boundary of traffic signs using convolutional neural network (CNN). Estimating the precise boundary of traffic signs is important in navigation systems for intelligent vehicles where traffic signs can be used as 3D landmarks for road environment. Previous traffic sign detection systems, including recent methods based on CNN, only provide bounding boxes of traffic signs as output, and thus requires additional processes such as contour estimation or image segmentation to obtain the precise sign boundary. In this work, the boundary estimation of traffic signs is formulated as a 2D pose and shape class prediction problem, and this is effectively solved by a single CNN. With the predicted 2D pose and the shape class of a target traffic sign in an input image, we estimate the actual boundary of the target sign by projecting the boundary of a corresponding template sign image into the input image plane. By formulating the boundary estimation problem as a CNN-based pose and shape prediction task, our method is end-to-end trainable, and more robust to occlusion and small targets than other boundary estimation methods that rely on contour estimation or image segmentation. The proposed method with architectural optimization provides an accurate traffic sign boundary estimation which is also efficient in compute, showing a detection frame rate higher than 7 frames per second on low-power mobile platforms.
\end{abstract}


\IEEEpeerreviewmaketitle

\section{INTRODUCTION}
\IEEEPARstart{T}{raffic} sign detection has been a traditional problem for intelligent vehicles, especially as a preceding step for traffic sign recognition which provides useful information such as directions and alerts for autonomous driving or driver assistance systems. Recently, traffic sign detection has received another attention from navigation systems for intelligent vehicles, where traffic signs can be used as distinct landmarks for mapping and localization. Different from natural landmarks such as corner points or edges, traffic signs have standard appearance such as shapes, colors, and patterns defined by strict regulations. This forms a primary reason that traffic signs are a preferable choice as landmarks for high-definition map reconstruction, as it allows efficient and robust landmark detection and matching under various conditions.

To reconstruct detected traffic signs to a 3D map, one should have point-wise correspondences of boundary corners of the signs across multiple frames, and then compute the 3D coordinates of the boundary corners by triangulation using the camera pose and the internal parameters of the camera (see \cite{mapping} for details of the 3D mapping procedure). For accurate triangulation of 3D positions, it is required to estimate the boundary of signs with pixel-level accuracy. However, previous traffic sign detection systems do not meet this requirement as they only estimate the bounding box of a traffic sign. Pixel-wise prediction methods such as semantic image segmentation, which have been applied successfully for road scenes~\cite{segnet,seg_Uhrig,seg_Lin}, can be an alternative for boundary estimation. However, it requires time-consuming algorithms that can severely degrade the performance of real-time systems.

\begin{figure}[t]
\begin{center}
   \includegraphics[width=\linewidth]{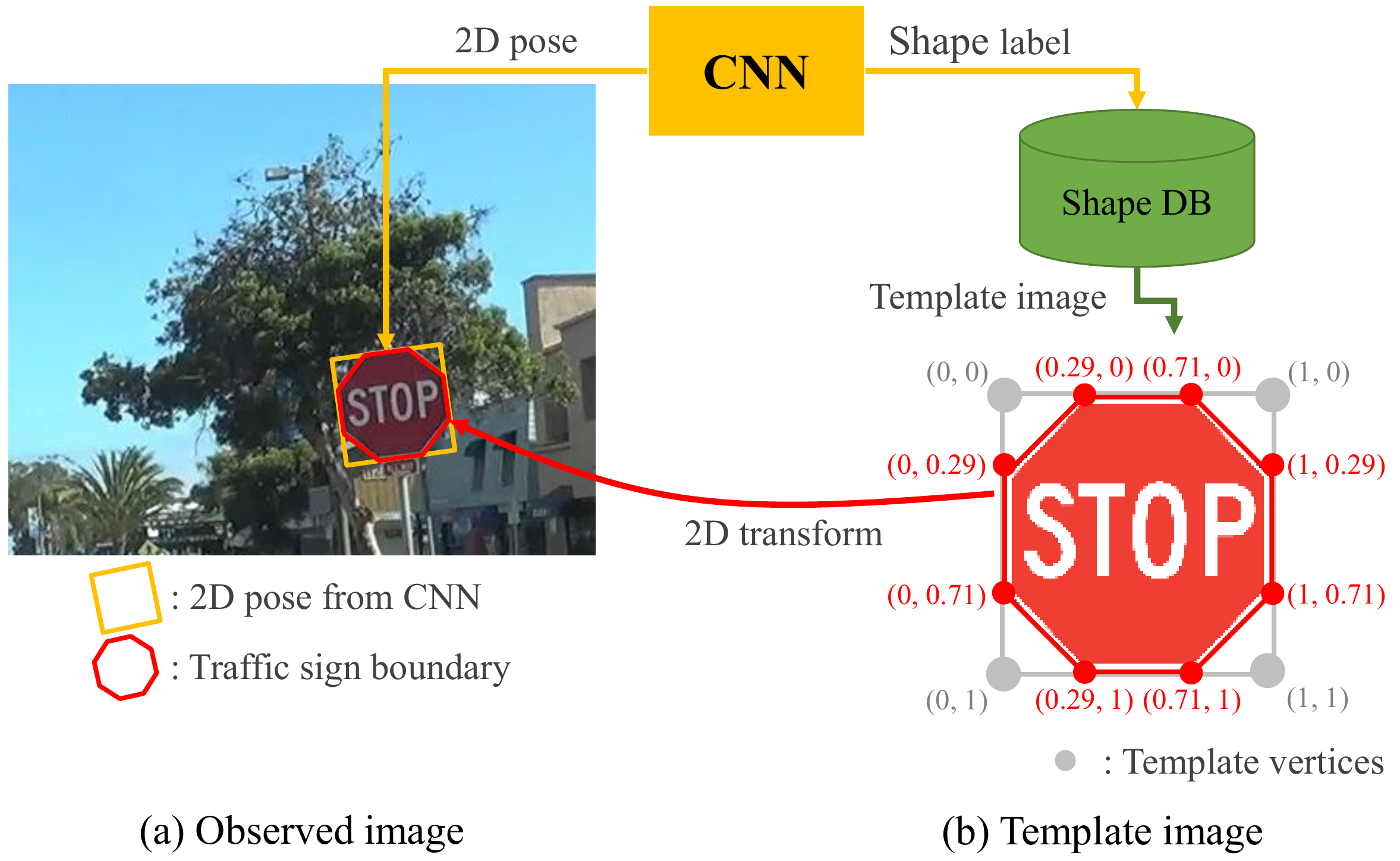}
\end{center}
\caption{Boundary estimation using a predicted 2D pose and a shape label: 2D poses and shape labels are predicted from CNN. The template image (b) of the predicted shape label is retrieved from the shape database, and then the boundary corners of the template image are projected to the observed image (a) by the predicted pose.}
\label{fig:1}
\end{figure}

To overcome these limitations, we propose a traffic sign detection system where the position and precise boundary of traffic signs are predicted simultaneously using a single convolutional neural network (CNN). Our novel object detection network is based on the recent advances in CNN-based object detection for object bounding box prediction~\cite{RCNN, SSD, FRCNN}, but tailored to predict the 2D poses and shape labels of planar targets. The 2D pose of a planar target can be encoded as an 8-dimensional vector, \eg the coordinates of four vertices, and it can be accurately predicted by CNN which simultaneously predicts the score of each shape label. Using the predicted 2D poses and shape labels, the boundary corners of a traffic sign are computed by projecting the boundary corners of a corresponding template image of the sign into the image coordinate using the predicted pose, as illustrated in Figure~\ref{fig:1}. 

By using the templates of traffic signs, our method effectively utilizes strong prior information of target shapes. This enables robust boundary estimation for traffic signs that are occluded or blurry, which is often challenging in pixel-wise prediction such as contour estimation and segmentation. As a result, our method achieves detection rates higher than 0.88 mean average precision (mAP), and boundary estimation error less than 3 pixels with respect to input resolution of 1280 $\times$ 720 pixels. Since projecting boundary corners (matrix-vector products) requires negligible computation time, most of the required computation is from the forward propagation of CNN which can be accelerated by GPUs. Combining with our efforts to find a base network architecture that provides the best trade-off between accuracy and speed, our precise boundary detection system can be run on mobile platforms with frame rates higher than 7 frames per second (FPS) with affordable traffic sign detection and boundary estimation accuracy.

The rest of paper is organized as follows. Section \ref{sec:related} reviews previous works on traffic sign detection as well as CNN-based generic object detection. The details of our method are described in Section \ref{sec:proposed} including the structure of our detection network, traffic sign boundary estimation using the network output, and training details. In Section \ref{sec:exp}, we report evaluation results on the proposed method including accuracy and speed, and intensive experiments on speed up of the network. Conclusion and future direction are summarized at the final section.

\section{RELATED WORK}\label{sec:related}

\subsection{Traffic Sign Detection}
Most of the previous work on traffic sign detection relies on handcrafted image features to identify target signs. Various combinations of features and classifiers are proposed to pursue a robust and fast traffic sign detector, and near one-hundred percent accuracy is achieved on small benchmark dataset such as German Traffic Sign Detection Benchmark (GTSDB) dataset~\cite{GTSDB}. Since running a complex feature extractor and a classifier is time-consuming, multi-stage cascade architectures composed of fast candidate search and accurate candidate classification have been a typical pipeline of traffic sign detection. 

For the early stage of the detection pipeline, simple but effective features such as color and edge are used to extract a number of candidate regions. Color is one of the most distinctive features of traffic signs, thus many works have utilized color features for fast candidate region search in various forms, such as color probability maps~\cite{Yang_2016,color_prob} and multiple thresholding~\cite{tsdr_multiple}. More discriminative features like saliency, textures, and patterns are favored in the middle stages of the pipeline. In \cite{Wang_2015}, saliency features are computed in multiple image scales, and then each feature of different scales is tested in cascade classifiers. In \cite{tsd_occ} and \cite{MN_LBP}, variants of local binary pattern (LBP) are extracted and fed into cascade classifiers. Region-based features like histogram of gradient (HoG), which compute the statistics of primitive features in the region, are widely used in the final stage of the detection pipeline where an accurate decision on traffic sign/non-traffic sign is required \cite{Yang_2016, Wang_2015, Wang_2013}.

Meanwhile, more sophisticated features such as the integral channel features (ICF)~\cite{ICF} or the aggregated channel features (ACF)~\cite{ACF} have also been applied to traffic sign detection systems~\cite{TSD_US, TSDTR_ieee_ITS}. The ICF/ACF features have strong discriminative power while being efficiently computed, thus a single classifier shows competitive accuracy compared to the cascade classifiers with simpler features.

Depending on implementation, traffic sign detectors based on the handcrafted features can be very fast without any special hardware (\eg detection module in \cite{TSDTR_ieee_ITS} takes 76ms on PC without GPU). Their accuracy, however, could be severely affected by hyper-parameters and various conditions such as weather, light, or camera properties, as features designed by hand are known to be weaker than features learned by CNN from massive training data~\cite{Alex}.

While traffic sign recognition has utilized CNN earlier \cite{TSC_idsia} and shown highly accurate results, the use of CNN in traffic sign detection has started recently in \cite{Wu_2013, Qian_2015}. In these works, the use of CNN is simply to apply CNN classifier to candidates regions to reject non traffic sign candidates, thus their accuracy and efficiency are significantly affected by candidate extraction step which still relies on handcrafted features. More comprehensive reviews and comparisons on the recent advances in traffic sign detection and recognition systems are studied in \cite{tsdr_review,tsrd_review2,tsrd_review3}.

\subsection{CNN for Object Detection}
Recently, great advances have been achieved on object detection by CNN~\cite{RCNN, SSD, FRCNN}. Besides the discriminative power of CNN on object category classification, the detection networks show the capability of accurate localization by performing \textit{regression} on object location (this location regression will be explained at Section~\ref{sec:proposed}). Two different architectures of detection networks are currently being developed: direct detection~\cite{SSD,yolo} and region proposal based detection~\cite{FRCNN,rfcn}. In direct detection, predictions on position (by regression) and class (by classification) of target objects are directly obtained from convolutional feature layers, resulting in relatively faster run time. On the other hand, region proposal based methods first generate a number of candidate regions regardless of their classes, and then predict object position and class for each candidate region. By performing regression and classification twice in different stages of the network pipeline, the region proposal based methods pursue more accurate detection while being relatively slower than direct detection methods~\cite{speed}. For the case of traffic sign detection for autonomous driving, direct detection methods are arguably more adequate since the latency of detection is important under limited computational resources.

Although most of the recent CNN object detection methods provide accurate bounding box and class label prediction, further processes should follow to obtain precise object boundaries from the predicted bounding boxes. To resolve this issue, in \cite{TSD_wild}, boundaries of traffic signs are simultaneously obtained as segmentation masks by \textit{OverFeat}~\cite{overfeat}-style CNN trained on multiple tasks comprising bounding box detection, mask segmentation, and sign category classification. However, predicting pixel-wise segmentation masks requires intensive computation, resulting in very slow speed of the network. On the other hand, we propose boundary estimation method which does not require pixel-wise segmentation and thus enables faster detection speed.

\section{PROPOSED METHOD}
\label{sec:proposed}

\begin{figure*}[t]
\begin{center}
   \includegraphics[width=0.8\linewidth]{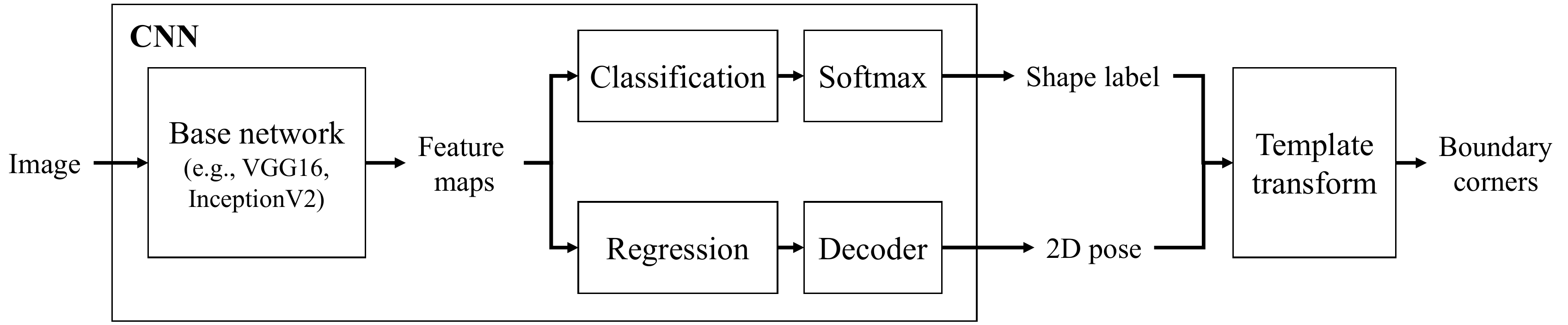}
\end{center}
\caption{The overall procedure of the proposed method. Inside CNN, convolutional feature maps are extracted from the input image, and the shape labels and 2D poses for each default box in a grid are predicted from the feature maps. Then we transform the corners of the template image that corresponds to the shape label using the 2D pose to obtain the boundary corners in the input.}
\label{fig:2}
\end{figure*}

The overall procedure of the proposed method is illustrated in Figure~\ref{fig:2}. For the CNN block, any object detection network structure~\cite{SSD,yolo,FRCNN,rfcn} which regresses the position of a target object can be used. In our work, we build the CNN block based on the SSD structure where predictions are directly preformed across multiple feature levels. The main difference of our network with the previous detection networks is what it predicts as output: instead of predicting bounding box coordinates, our network performs \textit{pose estimation}, which can be converted into the boundary estimation of corresponding traffic signs. In the CNN block, an input image is passed to a base network which extracts feature maps using a series of convolution, non-linear activation, and pooling operations. Then from the feature maps, 2D poses and shape class probabilities are estimated by two separated convolutional layers, namely, \textit{pose regression layer} and \textit{shape classification layer}, combined with successive operations that convert the convolution outputs to the 2D pose values and class probabilities, respectively (through Softmax and Decoder in Figure~\ref{fig:2}). Finally, we use the obtained 2D poses and shape class probabilities to compute boundary corners.

\subsection{Convolutional Feature Extraction}
Since pre-trained CNNs on ImageNet classification task~\cite{Alex,vgg,inception,Resnet} are popular starting points of fine-tuning for other vision tasks~\cite{decaf}, most of the recently introduced object detection methods also borrow one of these networks for feature extraction upon which detection layers perform bounding box estimation and object category classification. We also choose our base networks among these networks with a notable consideration: since we aim at using our method for real-time autonomous vehicle applications, the most important criterion to choose our base network is the accuracy/speed trade-off. For detailed comparison on base network for object detection, refer to \cite{base_comp}, and we present experimental results using VGG16 and InceptionV2 base network in Section~\ref{sec:exp}.

Similar with other CNN based object detection methods, our pose regression and shape classification layers perform on top of convolutional feature maps generated from these base networks. Since our implementation adopts SSD structure where detection is performed on multiple feature maps of different spatial resolutions, our pose regression and shape classification layers also spread in multiple layers to reliably detect traffic signs of different scales.

\begin{figure}[t]
\begin{center}
   \includegraphics[width=\linewidth]{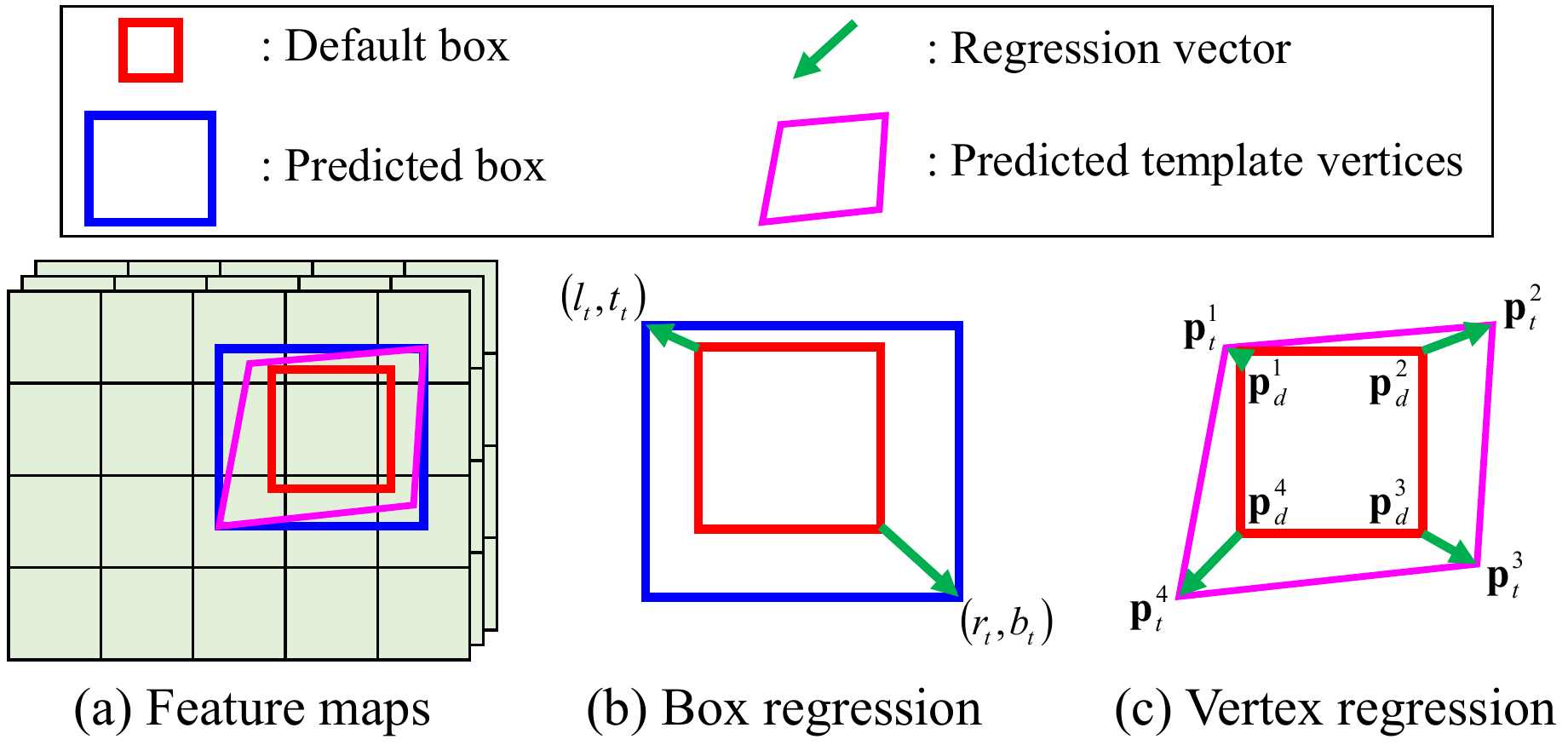}
\end{center}
\caption{Box and pose prediction using regression vectors. (a) For every unit of feature maps in the regression layer, we assign default boxes and obtain regression vectors for each of them. (b) In conventional object detection networks, two regression vectors are predicted for bounding box estimation. (c) In our model, four regression vectors are predicted for 2D pose estimation.}
\label{fig:3}
\end{figure}

\subsection{2D Pose Prediction}
We first review the bounding box regression used in recent CNN-based object detectors~\cite{SSD,FRCNN}. The regression layers, which can be either convolution layers or fully-connected layers, take feature maps from the base network and predicts pose offsets relative to the default box coordinates. Here, a set of default boxes of different aspect ratios and scales is assigned to each unit of the feature maps in the regression layers as shown in Figure~\ref{fig:3}(a). Since a bounding box can be represented by two 2D coordinates, \ie (left, top, right, bottom), 4-dimensional vectors can be used to represent regression values. For example, the regression vector can be represented as $(\Delta l, \Delta t, \Delta r, \Delta b) = (l_t, t_t, r_t, b_t) - (l_d, t_d, r_d, b_d)$, where $l, t, r$, and $b$ are the left, top, right, and bottom coordinate of a bounding box, respectively, and the subscripts $_t$ and $_d$ denote `target' and  `default'. A target box corresponds to a ground truth box in the training stage and a predicted box in the inference stage. In Figure~\ref{fig:3}(b), two green arrows correspond to $(\Delta l, \Delta t)$ and $(\Delta r, \Delta b)$, and by adding these two vectors to left-top and right-bottom corners of default box respectively, a bounding box can be obtained as a detection result.

In our method, on the other hand, the box regression is generalized to vertex regression to predict the 2D pose of targets. The 2D pose of a planar target in an image can be represented as perspective transform of the target, and the simplest way to determine the 2D pose is to obtain four points on the target plane. We predict an 8-dimensional regression vector of four points to obtain the 2D pose of the target. The four points need not be the actual corner points of the target, and arbitrary predefined locations on the target are enough to determine the 2D pose. To estimate the precise boundary of traffic signs, we use \textit{`template vertices'} of traffic signs (see Figure~\ref{fig:1}), which are virtual corners determining the minimum bounding quadrilateral of the signs as regression points. Then the regression vector can be defined as
\begin{equation}
	\Delta \bfp^i = \bfp_t^i - \bfp_d^i, i=1,2,3,4,
\end{equation}
where $\Delta \bfp^i$ is $i$th regression vector corresponds to $i$th target template vertex $\bfp_t^i$ and the vertex of default box $\bfp_d^i$, as illustrated in Figure~\ref{fig:3}(c). In the inference stage we can reversely estimate template vertices from the network output $\Delta \bfp^i$ as
\begin{equation}
	\bfp_t^i = \bfp_d^i + \Delta \bfp^i, \quad i=1,2,3,4.
\end{equation}

\begin{figure}[t]
\begin{center}
   \includegraphics[width=\linewidth]{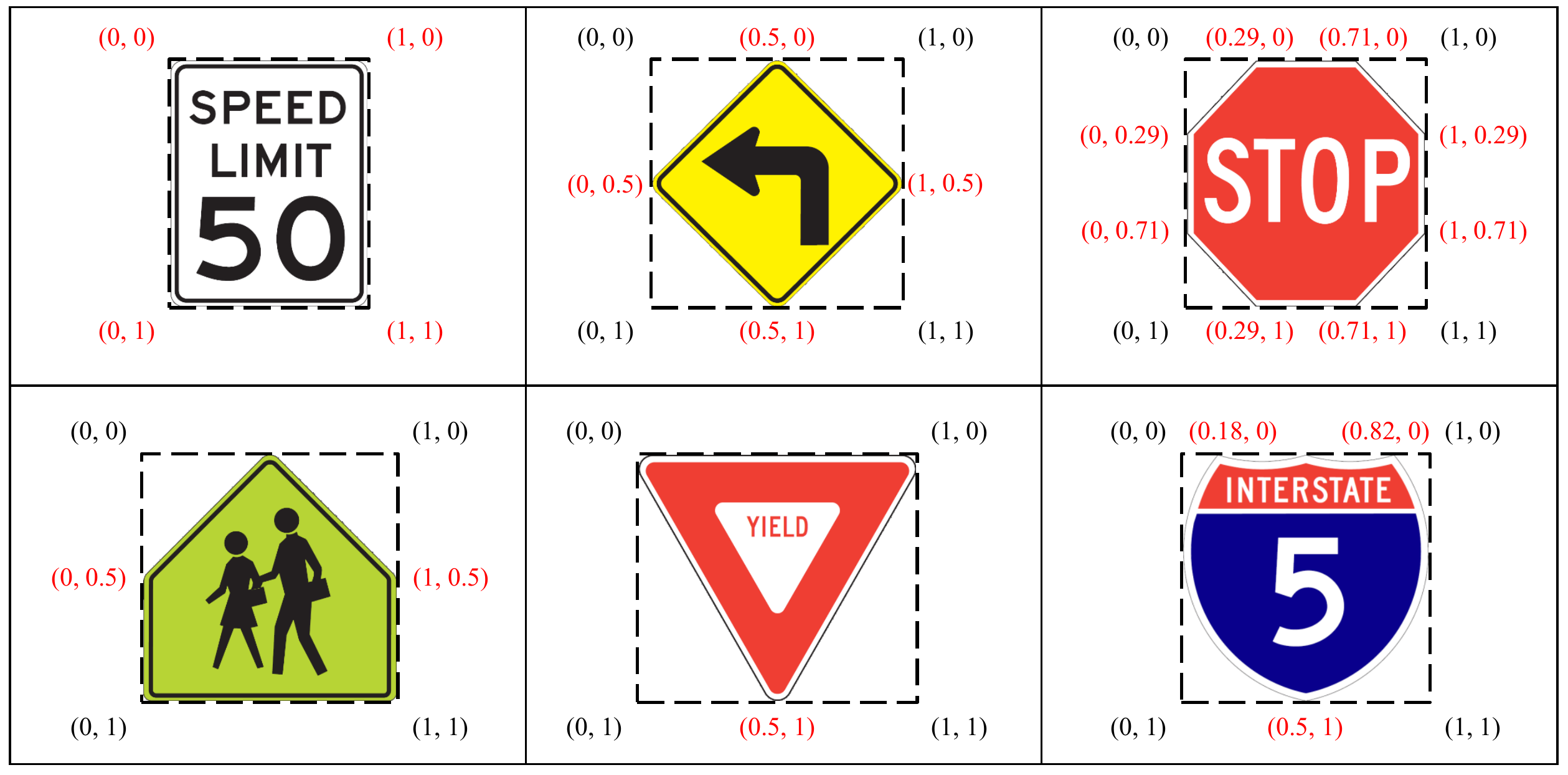}
\end{center}
\caption{Examples of traffic sign template images and their boundary corners.}
\label{fig:5}
\end{figure}

\subsection{Boundary Corner Estimation}
Given the 2D pose of a target sign, it is straightforward to determine the boundary of the sign given its template image. First, from the shape classification layer we predict the shape label of the target to determine which template image to use. The output of the shape classification layer, denoted by $\mathbf{s}$, is $N$-way softmax values where $N$ is the number of shape classes (rectangle, diamond, octagon and background in our experiments). We use one template image for each corresponding shape (for example, the shape label is octagon in Figure~\ref{fig:1}) where the coordinates of its boundary corners are already known as in Figure~\ref{fig:5}. For convenience, we normalize the coordinates of a template image, \ie the width and the height of the template image, to be 1. From the predicted template vertices $\{\bfp_t^i\}_{i=1,2,3,4}$, we can compute a 2D transform matrix $\mathbf{H}$ satisfying the following equation,
\begin{equation}
	\tilde{\bfp}_t^i = \mathbf{H} \tilde{\bfq}^i, \quad i=1,2,3,4,
\end{equation}
where $\tilde{\bfp}^i$ and $\tilde{\bfq}^i$ are the homogeneous representations of $\bfp^i$ and $\bfq^i$, respectively, and $\bfq^i$s are the coordinates of template image corners, such that $\bfq^1=[0, 0], \bfq^2=[1, 0], \bfq^3=[1, 1], \bfq^4=[0, 1]$. Finally, we transform the boundary corners $\{\bfc^j\}_{j=1,2,...,M}$ where $M$ is the number of boundary corners in the template image (\eg $M=8$ for octagon sign) using the computed transform matrix $\mathbf{H}$ into the image coordinate to obtain the precise boundary $\{\bfb^j\}_{j=1,2,...,M}$ of the detected traffic sign.

\begin{figure}[t]
\begin{center}
   \includegraphics[width=0.95\linewidth]{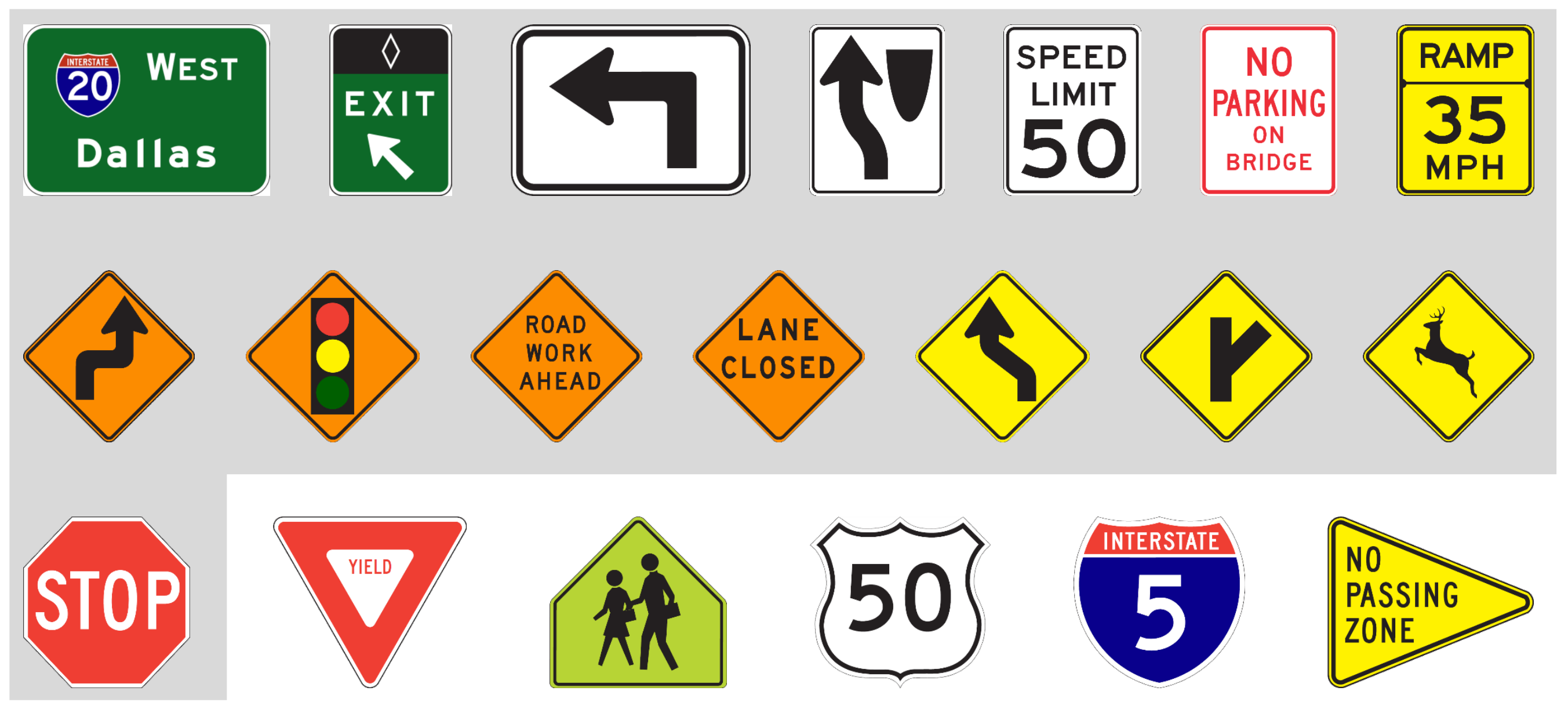}
\end{center}
\caption{Examples of U.S. traffic signs categorized by shape. \textit{Top row}: rectangle shape, \textit{Middle row}: diamond shape, \textit{Bottom row}: other shapes. Traffic sign shapes in the shaded area are used in our experiment.}
\label{fig:10}
\end{figure}

Figure~\ref{fig:5} shows examples of template images with various traffic sign shapes. The boundary corners $\{c^i\}$ are indicated by red coordinates. In case of `triangle' and `shield' shapes, we can represent the transform matrix $\mathbf{H}$ with an affine matrix, while other shapes having more than 3 corners can be represented by a perspective matrix. For traffic signs with circle shape, which are common in non-U.S. traffic signs, we can use four endpoints of major and minor axes of observed ellipse to compute the perspective transform matrix. In our experiments we only use rectangle, diamond, and octagon shapes since they are most frequent shapes in U.S. traffic signs. In fact, more than 90\% of U.S. traffic sign types have rectangle or diamond shapes~\cite{MUTCD}. Figure~\ref{fig:10} shows examples of traffic signs from \textit{California Manual on Uniform Traffic Control Devices}~\cite{MUTCD} for each shape class.

\begin{table}
    \centering
	\caption {Number of signs for each shape in \textit{`SDTS'} training and test sets}
\begin{tabular}[b]{|l||r|r|r|}
\hline
& Training set & Test set & Total\\
\hline
Number of images & 33,360 & 3,719 & 37,079\\
\hline
Number of signs & 57,692 & 6,324 & 64,016\\
Rectangle       & 45,961 & 5,188 & 51,149\\
Diamond         & 10,772 & 1,093 & 11,865\\
Octagon         & 959    & 43    & 1,002\\
\hline
\end{tabular}
\label{table:dataset_stat}
\end{table}

\subsection{Training}\label{sec:training}
All our models are based on base networks that are pre-trained on ImageNet dataset~\cite{imagenet}. The network parameters are fine-tuned during training the entire model. The pose regression and shape classification layers are trained from scratch with `Xavier' initialization~\cite{Xavier}. To train the pose regression and shape shape classification layers, we determine positive and negative samples and use them to compute a loss function. The positive samples correspond to predictions from default boxes matched with any of ground truth traffic sign, where `matched' means that the intersection over union (IoU) between the default box and the ground truth target is larger than 0.5. If a default box is not matched with any ground truth target, then the prediction from the default box is regarded as a negative sample.

We use the softmax loss as shape classification loss $L_{shape}(\mathbf{s})$ for all positive and negative samples, and use smooth $l1$ loss~\cite{fastrcnn} as vertex regression loss $L_{vertex}(\Delta \bfp^i)$ only for positive samples. For negative samples, the vertex regression loss is not assigned. The overall training loss for a single image is the sum of the shape classification loss $L_{shape}(\mathbf{s})$ and the vertex regression loss $L_{vertex}(\Delta \bfp)$ as follows,
\begin{equation}\label{eq:loss}
\begin{split}
	L_{overall}(\mathbf{s}, \Delta \bfp^i) = &\frac{1}{K_p+K_n} \lambda_s \sum^{K_p+K_n} L_{shape}(\mathbf{s}) + \\
	                           &\frac{1}{K_p} \lambda_v \sum^{K_p} \sum_{i=1}^{4} L_{vertex}(\Delta \bfp^i),
\end{split}							   
\end{equation}
where $K_p$ and $K_n$ are the number of positive and negative samples respectively, and $\lambda_s$ and $\lambda_v$ are loss weights for the shape classification loss and the vertex regression loss respectively. For notational simplicity, we omit the index of a default box inside each loss function. 

Since multiple default boxes of various aspect ratios are assigned to each unit of classification and regression layers, it is usual that the total number of default boxes is significantly larger than the number of positive samples, and most of the default boxes become negative samples. To resolve this severe imbalance between positive and negative samples, we employ \textit{hard negative mining} technique~\cite{SSD,hardnegative} to select only few of the negative samples with highest classification scores given by the classification layers. In our experiments, we set $K_n = 3K_p$ ($K_p$ is dependent on the number of ground truth objects in the image), and use only top-$K_n$ negative samples. We optimize this objective loss using the stochastic gradient descent method with initial learning rate $10^{-3}$, momentum 0.9 and batch size 32. The learning rate is decreased to $10^{-4}$ after 48,000 iterations, and optimization continues until 60,000 iterations.

\subsection{Dataset Manipulation}
For experiment data, we collected 37,079 road scenes containing traffic sings on highway as well as local road around San Diego, California, using a wide field of view camera mounted on a dashboard of a car. We refer to this `San Diego Traffic Sign' dataset as \textit{`SDTS'}, and the number of signs for each shape is reported at Table~\ref{table:dataset_stat}.\footnote{There are several benchmark dataset for traffic sign detection such as GTSDB dataset~\cite{GTSDB} for German traffic sign and LISA dataset~\cite{LISA} for U.S. traffic sign, but we can not use these datasets since they only have bounding box annotations.} The dataset is split into two disjoint sets; 33,360 training images and 3,719 test images which were captured at different paths. The collected images have slight radial distortion, but ground truth annotation, network training and testing are done without correcting the distortion. For each image, the boundary corners of traffic signs as well as their shape labels are annotated.

To train pose regression layers using the annotated data, we convert the annotated boundary corners $\{\bfb^j\}$ into template vertices $\{\bfp_t^i\}$ by calculating a transform matrix $\mathbf{H}$ from pairs of $\{\bfb^j\}$ and boundary corners $\{\bfc^j\}$ in the template image, and then transforming the image corners $\{\bfq^i\}$ of the template image using the matrix $\mathbf{H}$ to obtain $\{\bfp_t^i\}$.

\begin{figure}[t]
\begin{center}
   \includegraphics[width=0.95\linewidth]{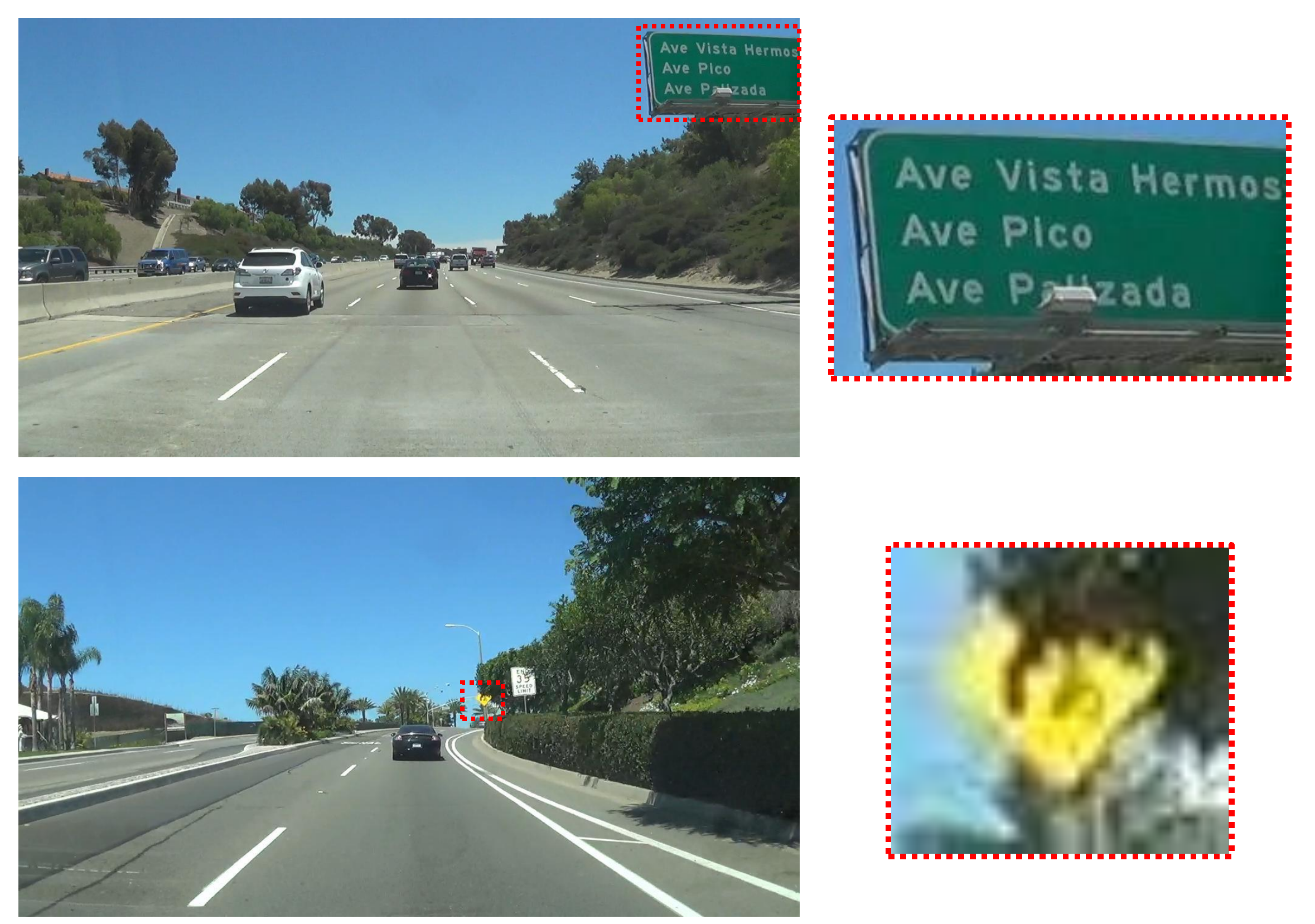}
\end{center}
\caption{Examples of unusable images for training. \textit{Top}: traffic sign touching image border. \textit{Bottom}: occluded traffic sign.}
\label{fig:7}
\end{figure}

For more effective training, we manipulate the training data with pruning, augmentation, and hard negative mining as follows.

\subsubsection{Pruning training data}
In case of occluded or partially observable traffic signs as shown in Figure~\ref{fig:7}, it is difficult to annotate accurate boundary corners of these signs. Since using the vertex regression loss from these samples may degrade the vertex prediction accuracy, we do not use them during training. In order not to modify the overall loss function (\ref{eq:loss}), we exclude images that contain at least one occluded or partially observable traffic signs, instead of not assigning vertex regression loss to those samples. Although those partially observable traffic signs are not used in training, the trained detector can robustly estimate boundary of such signs as shown in Figure~\ref{fig:13}.

\begin{figure}[t]
\begin{center}
   \includegraphics[width=\linewidth]{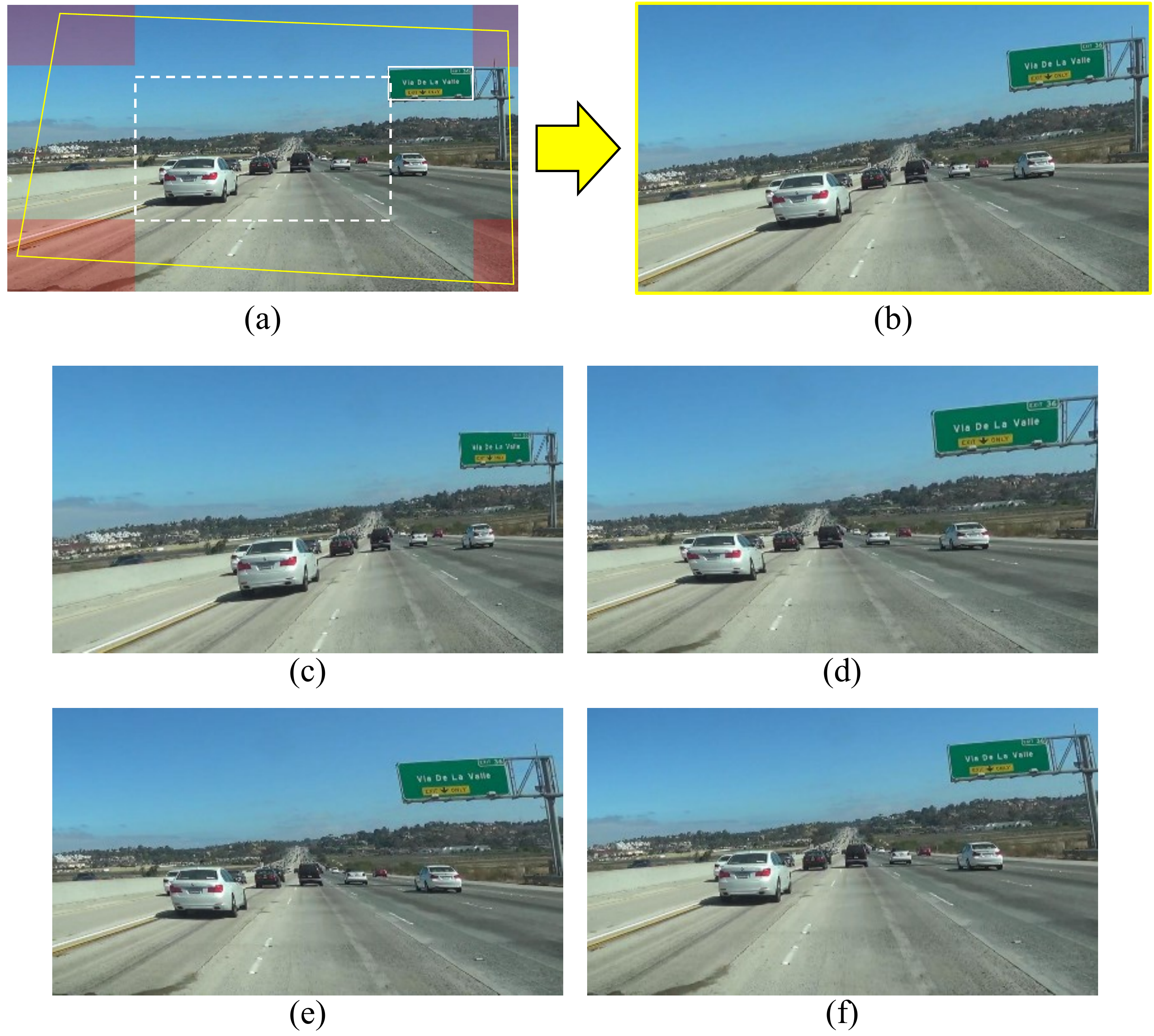}
\end{center}
\caption{Perspective data augmentation for close and large sign targets. In the original image (a), four corner points are randomly sampled from each red regions. The red regions are determined by the union of half center region (dotted white rectangle) and sign target (solid white rectangle). The yellow quadrilateral determined by the four corner points are then transformed to the augmented image with the original image size as shown in (b). Images in (c)$\sim$(f) are examples of augmented images from the same image. We generate 10 augmented images from one original image.}
\label{fig:6}
\end{figure}

\subsubsection{Data augmentation} \label{sec:augmentation}
To enhance the robustness of our detector to various appearance changes of traffic signs, we populate the training data with data augmentation techniques. We applied image cropping and perspective transform to training data: image cropping helps the detector be less sensitive to the position and size of traffic signs, and perspective transform makes the detector robust to view point changes. Especially, as the camera gets closer to a traffic sign, the sign generally moves toward the image boarders and has severe perspective distortion compared to distant signs. However, the number of close sign samples are limited compared to distant sign samples, because objects shift faster as they come close to the camera. We compensate this imbalance of samples between distant and close traffic signs by augmenting close signs with perspective transform. Only for images containing at least one large sign (about 620 images out of 33,360 training images), we generate randomly transformed duplicates for each of the images using perspective transform as shown in Figure~\ref{fig:6}. If a transformed image has a sign at the image border, we discard it and regenerate a new transformed image so that they do not contain partially observable traffic signs. In our experiment, we generated 10 augmented images from each original image, resulting in 6,200 augmented images in total.

\subsubsection{Hard negative mining}
In addition to image-level hard negative mining described in Section~\ref{sec:training}, we employ dataset-level hard negative mining in an aggressive manner to make our detector more robust to false positives. To our training data we add additional images that do not contain any positive samples and have confusing objects of traffic signs. This is a notable difference compared to previous works~\cite{SSD,FRCNN} in that they only use training images that contain at least one positive example. Concretely, we first train our model on training images all of which include traffic signs, and then run the trained model on images without traffic signs to find hard negatives. We sort these hard negatives and add the top 5,000 images to the training data in addition to the initial 33,360 images. Our final models used in our experiments were trained on this extended training data. Since traffic signs and frequently confusing objects (advertisement signs, some types of trees, etc.) are not often collocated, this strategy helps our models benefit from hard negative mining effectively. We confirmed in our experiments that adding these negative images significantly reduce false positives.



\section{EXPERIMENTS}
\label{sec:exp}

\subsection{Network Implementation}
The proposed pose regression layer can be combined with most of recent object detection networks such as Faster R-CNN~\cite{FRCNN} and SSD~\cite{SSD}. We particularly implement our models on top of the SSD framework since it is significantly superior to Faster R-CNN in terms of computational efficiency while being competitive in accuracy. As reviewed in Section~\ref{sec:related}, SSD generates position and class predictions directly from feature layers instead of using intermediate region proposals. To cover multiple scales of target objects, predictions of SSD are obtained from multiple feature layers of different resolutions. We use 6 feature layers of different resolutions (1/8, 1/16, 1/32, 1/64, 1/128, and 1/256 size of the network input) as detection sources, and we select the last layer among layers of the same resolution in the base network as the source layer of that resolution. The default boxes are defined with different scales and aspect ratios (1, 2, 3, 1/2 and 1/3 in our experiments) for each source layer, and redundant predictions from all the source layers are filtered out by non-maximum suppression (NMS) according to classification confidence.

\begin{figure}[t]
\begin{center}
   \includegraphics[width=\linewidth]{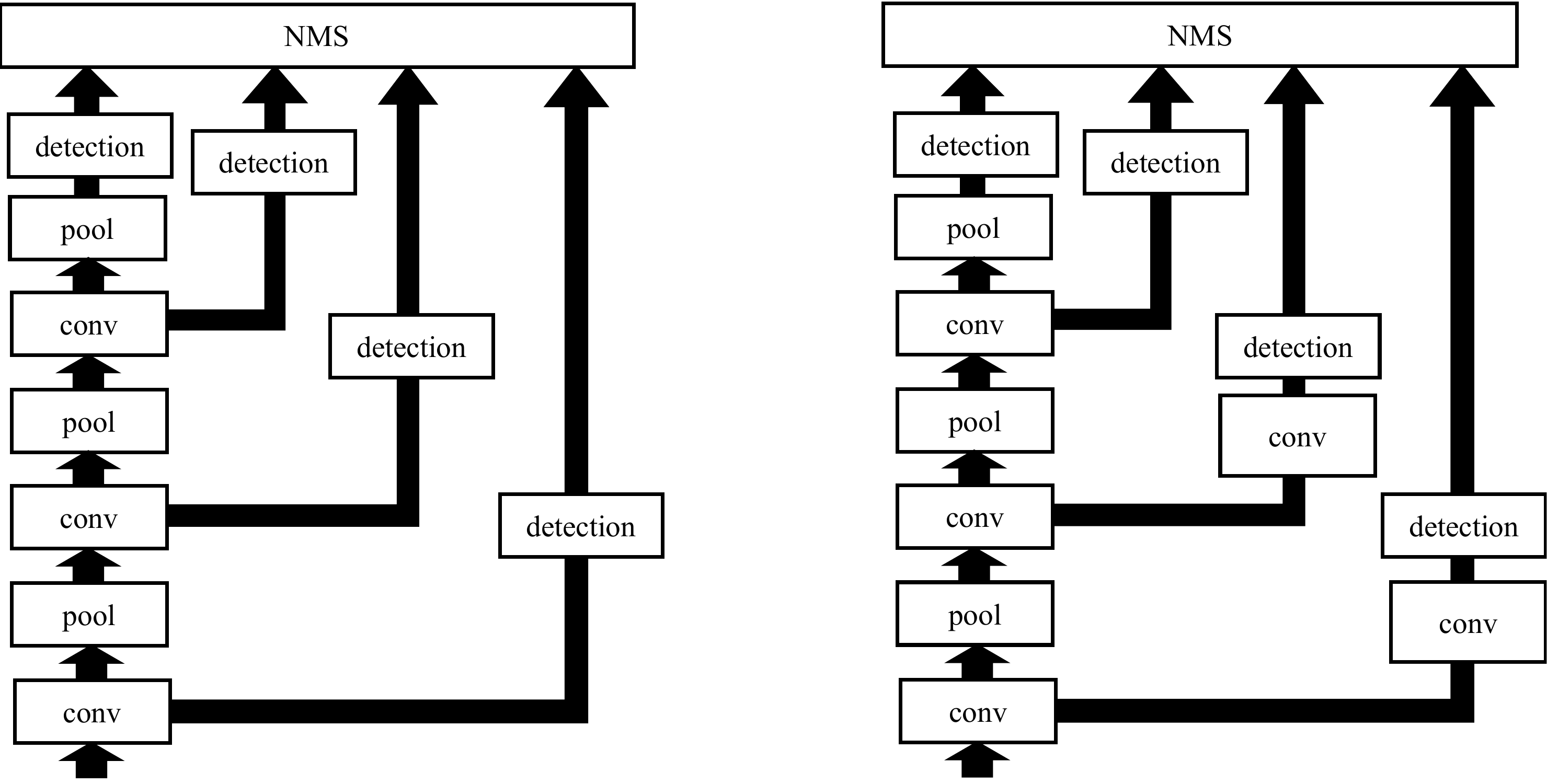}
\end{center}
\caption{Inserting additional convolutional layers into lower detection paths: model diagrams without (left) and with (right) additional convolutional layers.}
\label{fig:ssd_ext}
\end{figure}

Upon this SSD structure, we modify the network structure to strengthen the predictions in lower layers: we insert additional convolutional layers before the pose regression/shape classification layers, to compensate their relatively shallower depths from input. A graphical depiction of this approach is in Figure~\ref{fig:ssd_ext}. We apply this extension for two lower prediction layers (1/8 and 1/16 scales) since higher layers already have enough depths and receptive field sizes. Given the fact that small traffic signs appear more frequently (refer to section~\ref{sec:augmentation}), this extension is helpful to improve detection accuracy as discussed in section~\ref{sec:modelanalysis}.

\subsection{Experimental setup}
All the training process was conducted using \textit{Caffe} framework~\cite{caffe}. For testing, we implemented the system using Snapdragon Neural Processing Engine (SNPE) SDK which provides an optimized implementation of CNN forward propagation on the Qualcomm~\textsuperscript{\textregistered} Snapdragon~\textsuperscript{TM} 820A processor\footnote{Qualcomm Snapdragon~\textsuperscript{TM} is a product of Qualcomm Technologies, Inc.} with GPU utilization. For both training and testing, we used RGB channel color images, and different image resolutions for base network are tested to evaluate trade-off between accuracy and speed.

\subsection{Accuracy Evaluation}
Before presenting the evaluation of speed optimized network (Section~\ref{subsec:optimized_net}), we first present the accuracy evaluation of our detection network using VGG16 as a base network.

\subsubsection{Evaluation protocol}
We computed two measures for traffic sign detection accuracy: the mean average precision (mAP) and the average vertex error (AVE). The mAP is to measure detection rates and the AVE is to measure the accuracy of boundary estimation of detected signs. For each boundary estimation that matches a ground truth boundary (``match" corresponds to the condition that IoU between a detected box and a ground truth box is above a predefined threshold), the vertex error $v\_err$ is computed by the average distance between the estimated boundary corners $b^j_e$ and the ground truth boundary corners $b^j_g$ using the following equation:
\begin{equation}
	v\_err = \frac{1}{M} \sum_{j=1}^M \| b^j_e - b^j_g \|,
\end{equation}
where $M$ is the number of corners in the traffic sign.

\subsubsection{\textit{`GTSDB'} test dataset}
Since our method requires polygon annotations for training as well as AVE evaluation, we cannot evaluate the boundary estimation accuracy of our method on existing benchmark datasets. However, it is possible to evaluate only the detection rate with respect to bounding box, thus we compare the accuracy of ours (trained on box annotations of training data) with the state-of-the-art methods on GTSDB test dataset as summarized in Table~\ref{table:GTSDB}. We can see that our detector using VGG16 base network gives the best accuracy with respect to average box overlap (AO) for all traffic sign categories, with competitive accuracy with respect to area under precision-recall curve (AUC).

\begin{table}
    \centering
	\caption {Bounding Box Accuracy Comparison on GTSDB (\%)}
\begin{tabular}[b]{|l||r|r|r|r|r|r|}
\hline
\multirow{2}{*}{Method} & \multicolumn{2}{|c|}{Prohibitive} & \multicolumn{2}{|c|}{Danger} & \multicolumn{2}{|c|}{Mandatory} \\
\cline{2-7}
& AUC & AO & AUC & AO & AUC & AO \\
\hline
wgy\@HIT501~\cite{Wang_2013}    & \textbf{100} & 90.08 & 99.91 & 86.39 & \textbf{100} & 79.41\\
visics~\cite{visics}            & \textbf{100} & 88.22 & \textbf{100} & 87.03 & 96.98 & 89.55\\
LITS1~\cite{LITS1}              & \textbf{100} & 86.91 & 98.85 & 86.34 & 92 & 89.49\\
BolognaCVLab\cite{BolognaCVLab} & 99.98 & 84.95  & 98.72 & 86.78 & 95.76 & 86.19\\
Ours                            & 99.89 & \textbf{91.93} & 99.93 & \textbf{91.87} & 99.16 & \textbf{91.47}\\
\hline
\end{tabular}
\label{table:GTSDB}
\end{table}

\subsubsection{\textit{`SDTS'} test dataset}

\begin{figure}[t]
\begin{center}
   \includegraphics[width=0.8\linewidth]{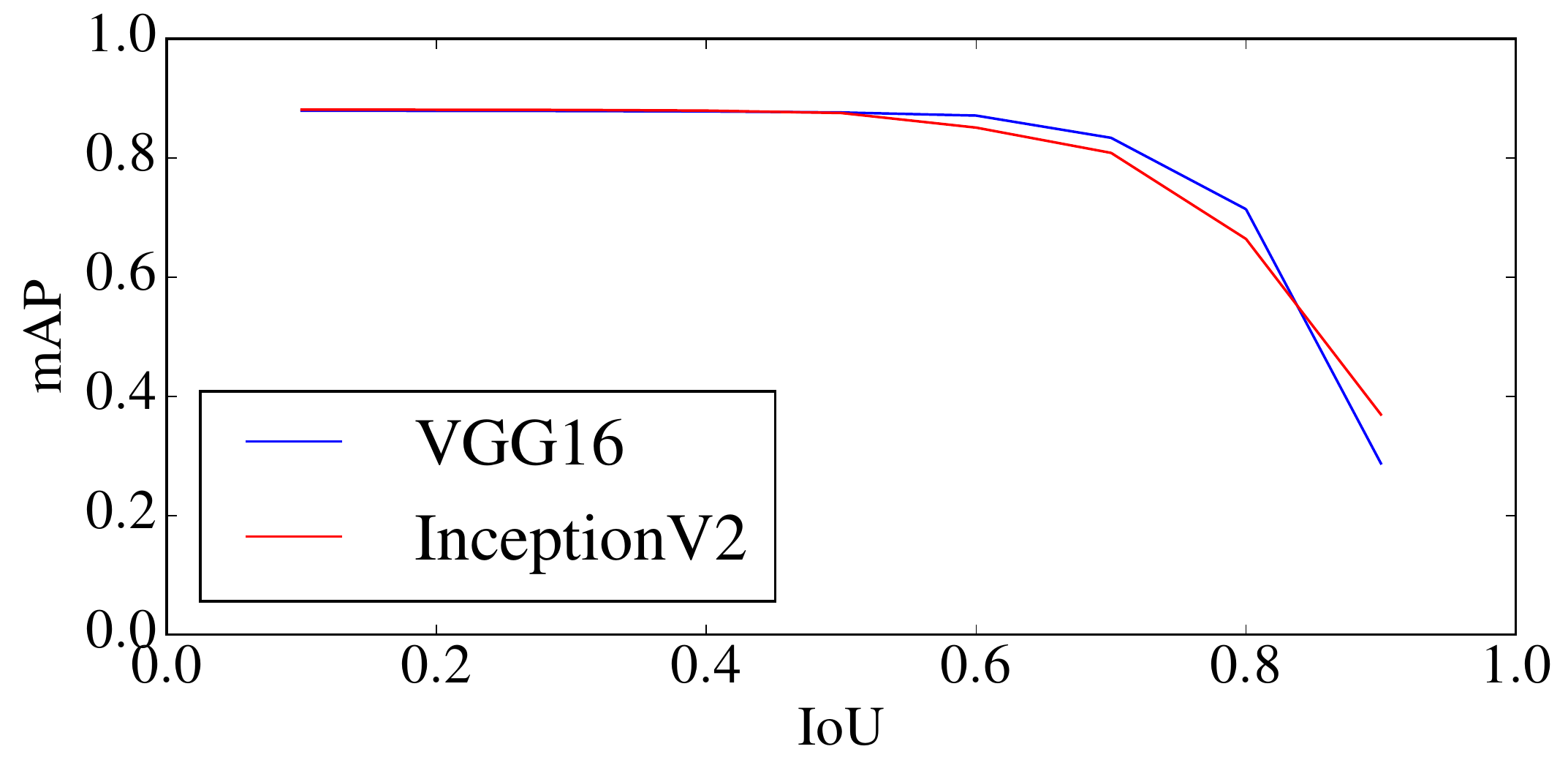}
\end{center}
\caption{mAP versus IoU plot by detectors based on VGG16 and InceptionV2, respectively. Our optimized InceptionV2-based detector shows comparable detection rate with VGG16-based detector.}
\label{fig:11}
\end{figure}

For accuracy evaluation, we used test images from \textit{`SDTS'} dataset which comprises 3,719 images with 6,324 traffic sign annotations. We calculate the mAP with varying IoU values, and the results are plotted in Figure~\ref{fig:11}. As shown in the figure, both models achieve mAP higher than 0.8 at 0.5 IoU, and even obtain mAP near 0.8 at 0.7 IoU. This shows that our methods are able to detect traffic signs with precise accuracy. More comparison between different base networks will be discussed in Section \ref{subsec:basenet}.

\begin{table}
    \centering
	\caption {Accuracy evaluation on traffic sign shape with VGG16 base network}
\begin{tabular}[b]{|l||r|r|r|r|}
\hline
Shape & \# GTs & Precision & Recall & AVE\\
\hline
\hline
Rectangle & 6444 & 0.895 & 0.705 & 2.590\\
Diamond   & 1124 & 0.887 & 0.877 & 2.089\\
Octagon   & 47   & 0.913 & 0.894 & 2.280\\
\hline 
Total     & 7615 & 0.894 & 0.732 & 2.499\\
\hline
\end{tabular}
\label{table:eval_shape}
\end{table}

As reported in Table~\ref{table:dataset_stat}, we have a very unbalanced number of training samples for each sign shape.
To check how does this data imbalance affects the detection accuracy, we measure accuracy numbers for each shape independently, and the results are reported at Table~\ref{table:eval_shape}. Interestingly, we can see that the accuracy of the rectangle shape is worse than other shapes although it has much more training samples. One plausible explanation might be the wide variety of appearances of the rectangular traffic signs, as well as the existence of many confusing objects of similar appearance such as advertisement signs.

\begin{figure}[t]
\begin{center}
   \includegraphics[width=\linewidth]{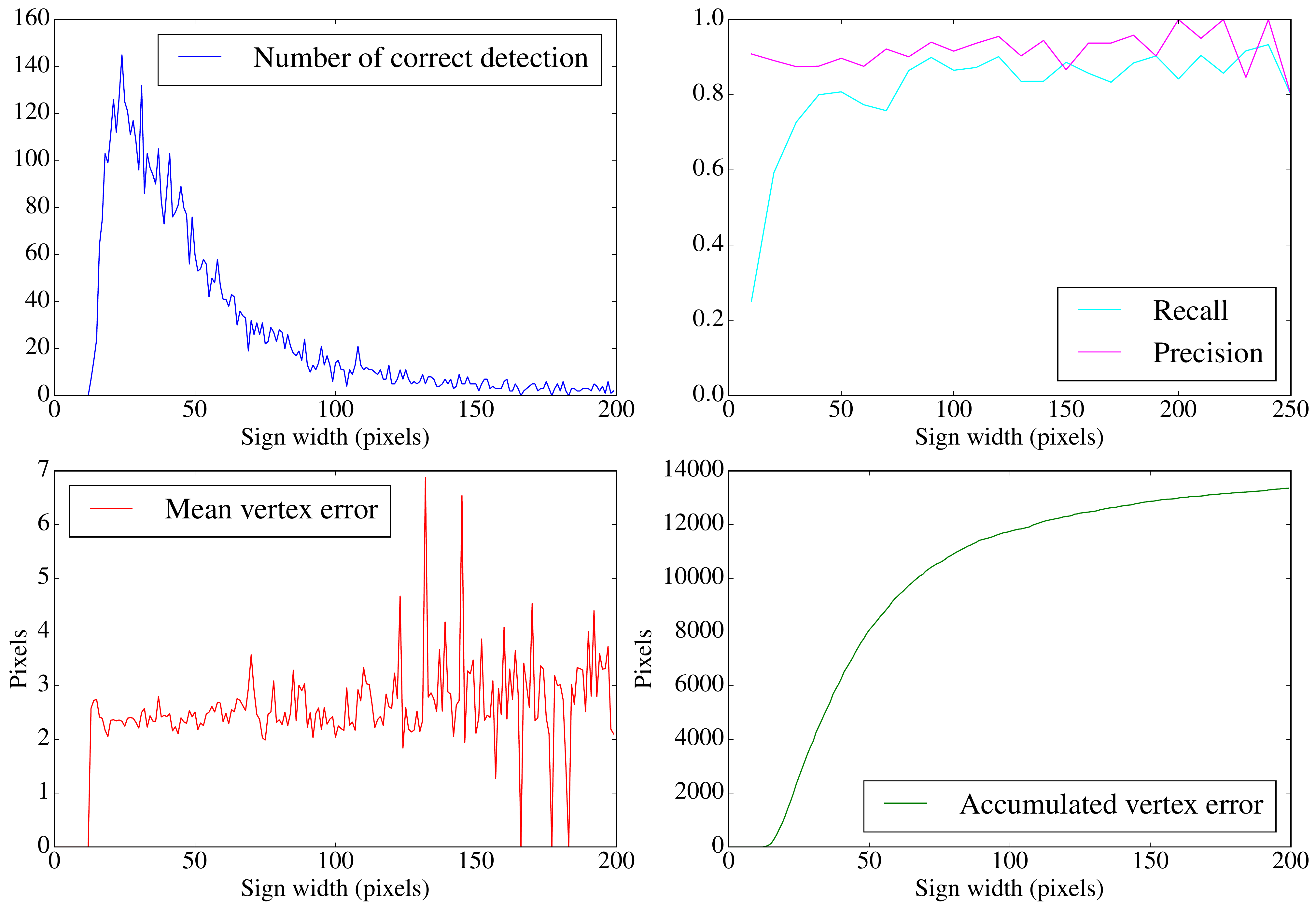}
\end{center}
\caption{Detection accuracy over traffic sign sizes. The size of signs is determined by the width of the sign in 1280 $\times$ 720 resolution.}
\label{fig:12}
\end{figure}

\begin{figure*}[t]
\begin{center}
   \includegraphics[width=\linewidth]{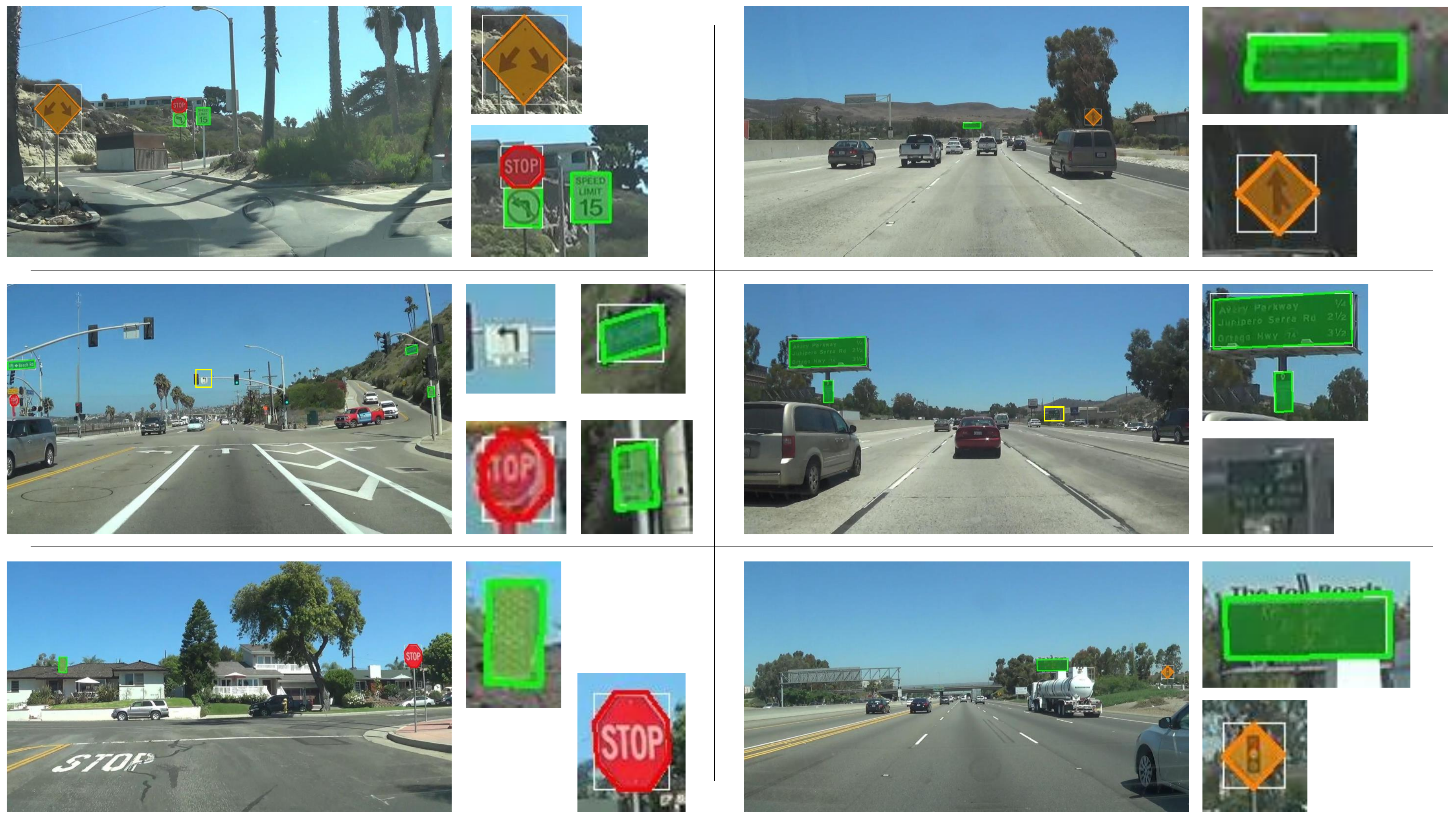}
\end{center}
\caption{Examples of detection results. Green, orange, and red colors indicate rectangle, diamond, and octagon shape, respectively, and white rectangles indicate bounding boxes. \textit{Left column}: images from local road scenes. \textit{Right column}: images from highway scenes. \textit{Top row}: All signs are detected correctly. \textit{Middle row}: Results include false negatives: small signs are not detected (locations of false negatives are indicated by yellow rectangles). \textit{Bottom row}: Results include false positives: a rectangular pattern is detected as a traffic sign in a local road scene, and an electronic sign board is detected in a highway scene.}
\label{fig:4}
\end{figure*}

Examples of detection results including failure cases are shown in Figure~\ref{fig:4}, and the video is available at the web \footnote{Qualcomm~\textsuperscript{\textregistered} drive data platform, \url{https://www.qualcomm.com/news/onq/2017/02/03/zoom-what-powers-qualcomms-drive-data-platform}}. In this experiment, input images to the base network have the resolution of $800 \times 450$ pixels, and signs whose side length is smaller than 13 pixels are ignored (Figure~\ref{fig:4} middle). As shown in the examples, our model is able to detect and estimate the precise boundaries of traffic signs with various sizes and shapes. Most of the false positives are caused by the objects of similar appearance, as shown in the bottom of the figure where a chimney or an electronic sign board is detected as a rectangular sign. With more training samples, especially various negative samples, these false positives can be reduced.

\subsubsection{Comparison with segmentation-based methods}
\begin{figure}[t]
\begin{center}
   \includegraphics[width=\linewidth]{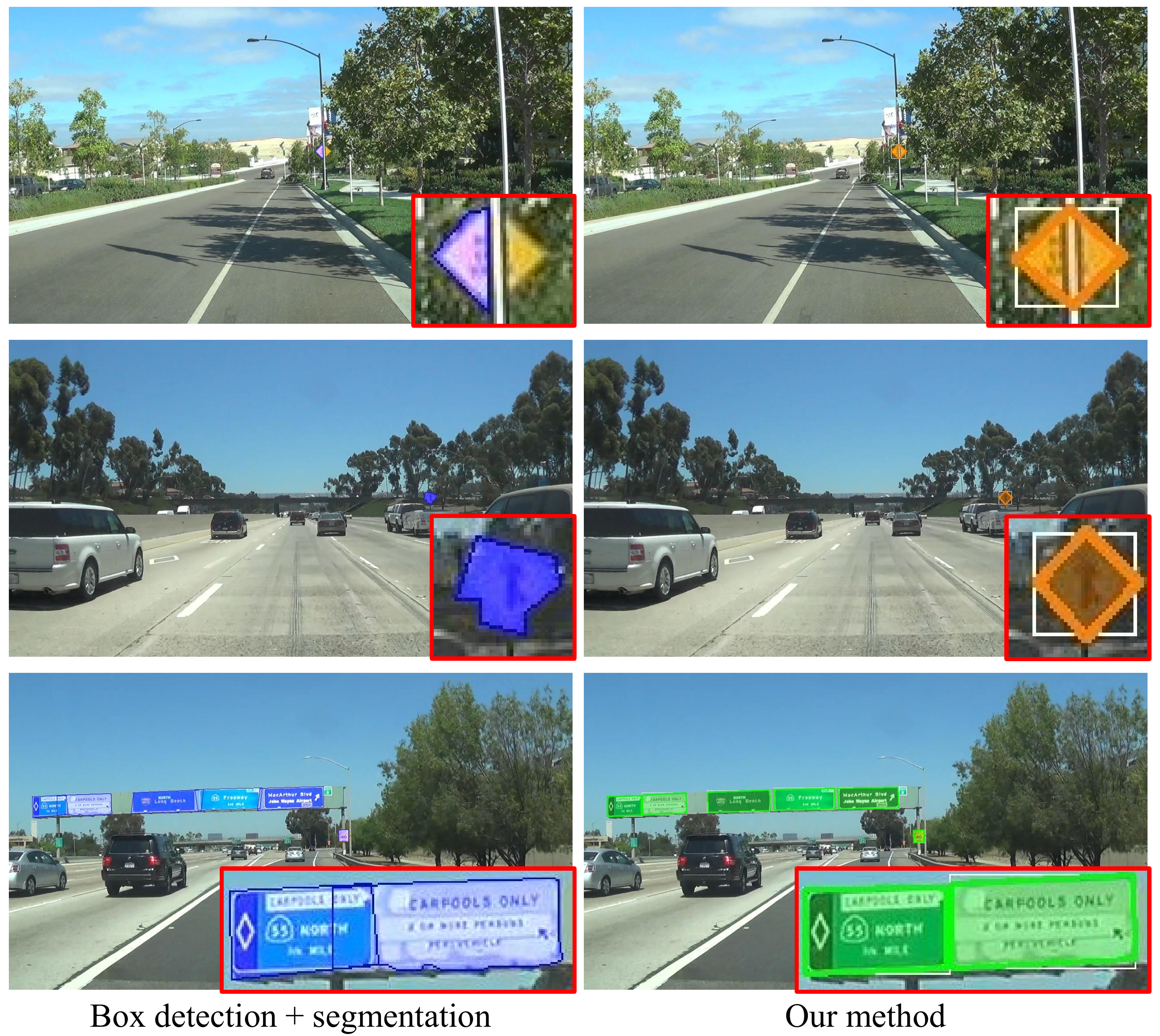}
\end{center}
\caption{Comparison of boundary estimation between our method and segmentation-based method. \textit{Top row}: occluded sign. \textit{Middle row}: cluttered background. \textit{Bottom row}: adjacent signs.}
\label{fig:13}
\end{figure}
We compare the boundary estimation accuracy of our method with a segmentation-based method combined with bounding box detection. For the segmentation-based method, same detection results are used with our method, but the results for the segmentation-based method are represented by bounding boxes (circumscribed rectangles of template vertices). For each bounding box, a patch is cropped with boundary padding, and then the patch is passed to the segmentation network (based on \cite{segnet}) trained on cropped traffic sign patches to generate binary masks. On the boundary of the estimated mask, polygon fitting and simplification are applied to obtain final boundary corner output. As shown in Figure~\ref{fig:13}, our method is more robust to partial occlusion, cluttered background, and adjacent signs which can not be directly handled by segmentation algorithm: since the segmentation-based method cannot restrict its output masks to fit to traffic sign shapes, an additional process for boundary fitting is required to refine the masks. We compare the AVE of the two methods: in VGG-16 based models, the segmentation-based method gives 3.2917 pixels error while our method gives 2.7014 pixels error.

For speed comparison, we run both methods on PC with \textit{Titan X} GPU. Our method takes 126ms per image, while the segmentation-based method takes 189ms per image excluding computation time for boundary fitting (106ms for SSD to detect bounding boxes and 83ms for the segmentation network). Depending on which boundary fitting algorithm to use, overall computation time of the segmentation-based method can be much increased.

\subsubsection{Sign boundary refinement}
Though our models provide robust boundary estimation for traffic signs, we observed that the accuracy can be improved further with an image-based local refinement by using the edge or gradient of sign boundaries. In the refinement process, we compute the gradient image for the patch of a detected traffic sign, then estimate an optimal 2D transform that minimizes the weighted distance between the computed gradient image and the predicted boundaries from our network. The 2D transform can be either perspective or affine, and in our experiments affine transform gives faster and more robust refinement results. Since the goal of this step is local refinements, we discard the refinement result if it is far from the original prediction given by the network. By this refinement, the average vertex error is reduced from 2.679 to 2.256, while it takes 1.2ms per image on PC.

\subsubsection{Volume of training data}
One drawback of CNN-based detection networks is that it requires a large number of training images. Although we prepared more than 44,000+ images (including 33,000+ images with traffic sign annotations, 5,000 negative-only images, and 6,000+ augmented images) for training, it is not clear that this amount of data is sufficient and more data can improve the accuracy. Therefore, we experimented with different number of training images and evaluated the accuracy of the trained networks. As reported at Table~\ref{table:training_size}, the detection accuracy marginally increases with larger data volumes, when the number of training images is larger than 15,000. The effect of data augmentation is reported at Table~\ref{table:model_analysis}.

\begin{table}
    \centering
	\caption {Detection accuracy by a different number of training images. VGG16 is used for the base network.}
\begin{tabular}[b]{|l||c|c|}
\hline
\# training images & mAP @0.5 IoU & AVE\\
\hline\hline
5,000  & 0.540 & 3.182\\
10,000 & 0.892 & 2.882\\
15,000 & 0.900 & 2.658\\
20,000 & 0.884 & 2.620\\
25,000 & 0.887 & 2.570\\
30,000 & 0.887 & 2.600\\
\hline
\end{tabular}
\label{table:training_size}
\end{table}

\subsection{Speed Optimized Network}\label{subsec:optimized_net}
Though our CNN based traffic sign detector shows reliable accuracy, the heavy computational demands from many layers hinder our models from running on low-power devices for autonomous driving and mapping. For example, our best-performing model relies on VGG16 network, which requires more than 100 billion multiply-accumulate operations (MAC) in a single forward pass for 800 $\times$ 450 resolution input. On GPU-supported Snapdragon~\textsuperscript{TM} 820A processor, this amount of computation only provides less than one FPS, surely far from a minimal requirement for our target applications. To achieve practical FPS, we apply three approaches all of which directly reduce the number of required MAC of our models: using lighter base networks, reducing the input image resolution and cropping input images.

\subsubsection{Lighter base network}\label{subsec:basenet}
Recent advances in deep CNN have shown that increasing the depth of neural networks increases accuracy~\cite{inception,Resnet}. However, the accuracy gain is relatively marginal compared to the increased computational cost, and this prevents adopting  recent deep CNN to low-power devices where computational capacity is clearly limited. To make our models accommodate to low-power devices, we replace the base network from VGG16 to a computationally lighter network. We select InceptionV2~\cite{InceptionV2} network as an alternative since it has shown to provide a good trade-off of accuracy and speed~\cite{base_comp}. InceptionV2 requires about 14 billion MAC in one forward pass for 800 $\times$ 450 resolution input, less than one seventh of MAC for VGG16.

Replacing the base network from VGG16 to InceptionV2 leads significant increase in FPS, as shown in Table~\ref{table:basenet}. Nonetheless, it just provides less than 3 FPS, still far from realizing our systems practical. To further speed up our network, we decrease MAC of InceptionV2 network by removing its several higher layers. The motivation of this approach is that our target objects (\ie traffic signs) are relatively simple in their shapes and structures compared to generic object classes, thus the higher layers have less importance since they are known to be needed to learn hierarchical structures of object parts and complex shapes. To verify this approach, we conducted experiments on using only some lower part of InceptionV2 network for our task, and Table \ref{table:basenet} shows the results. As expected, removing most of upper layers does not significantly affect detection accuracy. We found that removing all upper layers from `inception4b' module reduces almost half of MAC yet degrades detection accuracy marginally.

\subsubsection{Input image resolution}
Since our network is fully-convolutional, MAC required for our model are directly proportional to the number of input pixels. This means that the input image resolution is a key factor of overall computational cost. To reduce the input resolution while keeping detection accuracy as much as possible, we conducted experiments on the trade-off between speed controlled by input resolution and accuracy measures (mAP and AVE). Table ~\ref{table:resolution} shows that 533 $\times$ 300 input shows near 5 FPS with competitive accuracy to higher resolution settings, providing a reasonable trade-off between speed and accuracy. Sacrificing little more accuracy leads to more than 7 FPS by 400 $\times$ 225 resolution input. With a robust object tracking method, our traffic sign detection of around 5 FPS could be a reasonable option for near real-time traffic sign detection systems.

\begin{table}
    \centering
	\caption {Performance by different base network. MAC numbers are calculated based on $224\times224$ input resolution and for base networks only.  FPS numbers are calculated for $800\times450$ input }
\begin{tabular}[b]{|l||c|c|c|c|c|c|c|}
\hline
Base Network & MAC (million) & mAP & AVE & FPS \\
\hline\hline
VGG16 & 15470 & 88.4 & 2.746 & 0.2\\
InceptionV2 & 2018 & 86.8 & 2.928 & 1.5\\
InceptionV2 cut4b & 1270 & 86.4 & 3.051 & 2.1\\
InceptionV2 cut4a & 1145 & 86.5 & 3.126 & 2.4\\
InceptionV2 cut3c& 1025 & 85.2 & 3.212 & 2.5\\
SqueezeNet & 390 & 75.4 & 3.818  & 4.3\\
\hline
\end{tabular}
\label{table:basenet}
\end{table}

\begin{table}
    \centering
	\caption {Performance by different input resolution With `InceptionV2 cut4a' optimized model.}
\begin{tabular}[b]{|l||c|c|c|c|c|c|c|}
\hline
Input Resolution (\# pixels) & mAP & AVE & FPS \\
\hline\hline
$800\times450$ (360,000) & 86.5 & 3.126 & 2.4\\
$600\times360$ (201,600) & 85.4 & 3.4425 & 4.2\\
$533\times300$ (159,900) &  85.2 & 3.5824  & 5.3\\
$400\times255$ (90,000) &  82.8 & 3.7238  & 7.3\\
\hline
\end{tabular}
\label{table:resolution}
\end{table}

\begin{figure}[t]
\begin{center}
   \includegraphics[width=\linewidth]{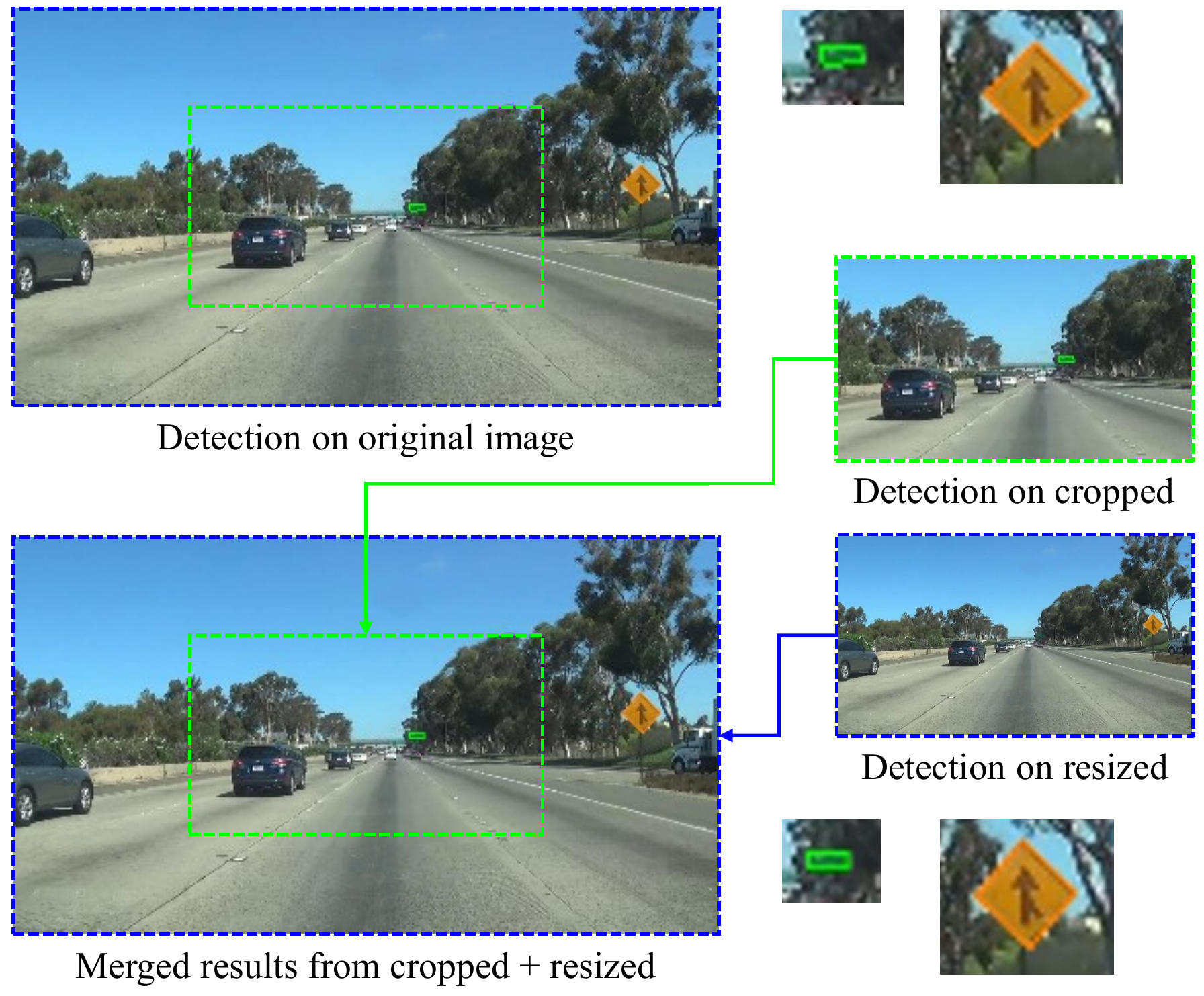}
\end{center}
\caption{Detection on cropped and resized images gives comparable results with detection on original images but with much less time.}
\label{fig:8}
\end{figure}

\subsubsection{Detection speed up by input cropping}
Besides the choice of base networks and input image resolution which are significant factors in accuracy/speed trade off, we applied a simple trick to improve the speed of our models by leveraging a characteristic of traffic scene. Different from generic object detection, there is a strong correlation between target sizes and positions for traffic scenes. Distant targets usually appear near the image center, and they are relatively difficult to detect due to their small sizes. On the other hand, close targets appear near the image border with larger sizes. Therefore, we can run two detectors with input images of difference sizes: one for a center-cropped image and the other for a half-resized image as shown in Figure~\ref{fig:8}. The center-cropped image has the same resolution with the original image and thus distant targets can be detected without accuracy degradation. On the other hand, close signs can be detected from the half-resized image with small accuracy degradation compared to the original image. Although we run two detectors, the total number of pixels becomes half than the original image, which results in half computation time for CNN compared to detection with the original image. In practice, including additional overhead such as image pre-processing and aggregating detection outputs from the two detector, the total running time of the two detector model is $2/3$ the time of the single detector model. The accuracy comparison is reported in Table~\ref{table:crop_resize}.

\begin{table}
    \centering
	\caption {Running two networks for cropped and resized images. \newline(time on PC with GPU, with VGG16 base network)}
\begin{tabular}[b]{|l||c|c|c|c|}
\hline
Model & Recall & Precision & AVE & time (ms)\\
\hline\hline
Original image  & 0.606 & 0.913 & 2.718 & 126.7\\
Crop \& resize  & 0.610 & 0.841 & 2.8839 & 79.0\\
\hline
\end{tabular}
\label{table:crop_resize}
\end{table}

\subsection{Model Analysis} \label{sec:modelanalysis}
\begin{table}
    \centering
    \caption {Effects of design choices and components on performance.}
    \setlength\tabcolsep{4pt}
\begin{tabular}[b]{|l|c|c|c|c|c|c|}
\hline
 & \multicolumn{6}{c|}{SSD InceptionV2 cut4a $533\times300$}\\
\hline
data augmentation &  & \checkmark & \checkmark & \checkmark & \checkmark & \checkmark\\
include {2, 1/2} boxes & \checkmark & \checkmark & \checkmark & \checkmark & \checkmark & \checkmark\\
include {3, 1/3} boxes &  &  &  & \checkmark & \checkmark \ & \checkmark\\
include {4, 1/4} boxes&   &   &  &  & \checkmark \ &\\
additional conv  &  &  & \checkmark & \checkmark & \checkmark \ &\\
additional  conv (light) &  &  &  &  &  & \checkmark \\
\hline
mAP &  83.3 & 84.4  & 85.3 & \textbf{86.5} & 86.2 & 86.3\\
AVE &  3.241 & 3.176  & 3.152 & 3.126 & \textbf{3.097} & 3.131\\
MAC (million) &  \textbf{5474} & \textbf{5474} & 6395 & 6472 & 6615 & 5905\\
\hline
\end{tabular}
\label{table:model_analysis}
\end{table}

We explore various SSD design choices to discover better architectures for our task. All the experiments are based on the `InceptionV2 cut4a' base network and 533 $\times$ 300 input resolution. Table~\ref{table:model_analysis} shows the results. The data augmentation by perspective transform improves accuracies on all model configurations, by 1.1\% of mAP. Choosing aspect ratios for default boxes affects accuracy, and the best model is obtained from five aspect ratios of 1, 2, 3, 1/2, and 1/3. Adding additional convolutional layers to increase the number of convolutional layers on the two lower detection paths also improves accuracy by 0.9\% of mAP. With the results in Table~\ref{table:basenet} and Table~\ref{table:resolution}, it can be seen that the vertex errors are more sensitive to input resolutions and base networks. The MAC of models varies marginally except the additional convolutional layers settings. This is caused by the number of feature maps in the additional layers (192 and 256, respectively). The increase of computational cost can be reduced by using a smaller number of feature maps in the added convolutional layers with marginal accuracy degradation as indicated in the right most column. The lighter layers have half the number of feature maps (96 and 128, respectively), reducing the MAC of the added convolutional layers significantly. 

\subsection{Mapping Accuracy} \label{sec:mapping}
\begin{figure}[t]
\begin{center}
   \includegraphics[width=0.9\linewidth]{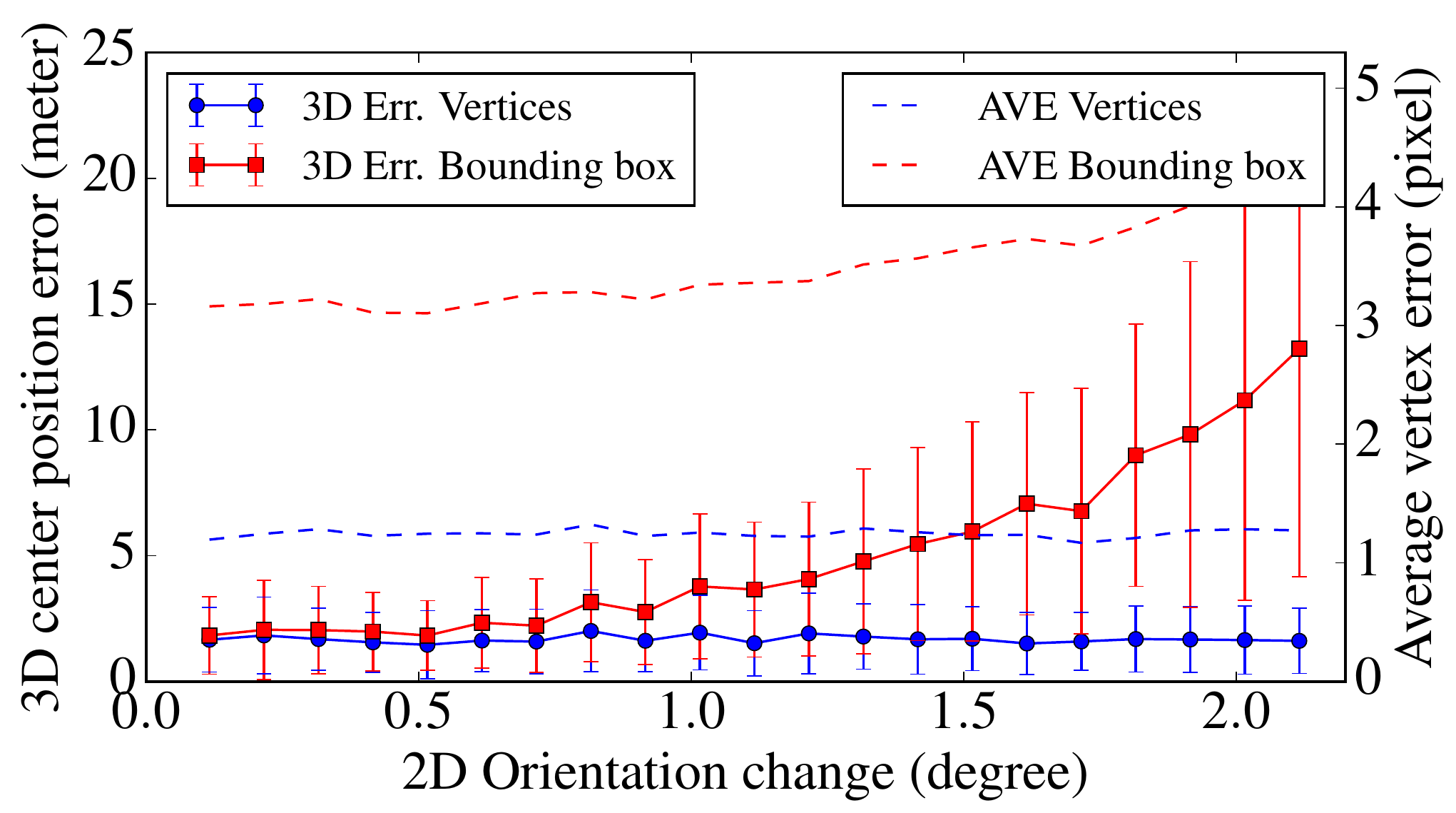}
\end{center}
\caption{Comparison of mapping accuracy using bounding boxes and vertices. \textit{Horizontal axis}: 2D sign orientation changes in degree across two frames. \textit{Vertical axis left}: errors of reconstructed 3D sign positions (3D Err.) using observations from two frames. The error plots are represented by averages and standard deviations from 100 simulation results with different observation noise. \textit{Vertical axis right}: Average vertex error (AVE) of the sign observed at the second frame. For bounding box detection, AVE is computed by the average distance between the ground truth vertices and their corresponding corners at the rectangular bounding box.}
\label{fig:14}
\end{figure}
Finally, we compare the traffic sign mapping accuracy between conventional bounding box detection and our precise boundary detection. If there is no geometric change of a sign (\eg rotation and perspective distortion) across multiple frames for mapping, then bounding box detection and our boundary detection will produce the same mapping accuracy. On the other hand, when we use bounding box detection instead of boundary prediction, even a small geometric change across frames can cause significant mapping error. Figure~\ref{fig:14} shows the simulated comparison of the mapping accuracy between bounding box detection and our boundary estimation, with varying 2D sign orientation changes induced by in-plane camera rotation (rotation around z-axis).

In this simulation, we synthesize images of $1280 \times 720$ resolution that emulate scenes from a forward-moving camera observing a rectangular sign. Concretely, we generate pairs of synthetic images each of which consists of two images of a sign observed from different camera positions. For all image pairs, the camera positions $(T_x, T_y, T_z)$ and $(T_x, T_y, T_z+\Delta)$ that simulate forward movement of distance $\Delta$ are fixed.\footnote{We set $T_x$, $T_y$, $T_z$ and $\Delta$ as $0$, $1$, $15$ and $5$ meters, respectively, and the focal length and principal point of the camera are set to 2000 and the image center, respectively. We set the center of the 3D sign to $(-7, 7, 25)$, and the size of the sign is set as $3 \times 2$ meters. The normal direction of the sign is set to be parallel with the camera z-axis.} The camera rotation around z-axis is 0 for the first frame and $\theta_z$ ($\theta_z \in [0, 2]$ degrees in Figure~\ref{fig:14}) for the second frame, while rotation around other axes remains constant, which results in 2D orientation change of sign between two frames. We project the four corners of the sign into the 2D image plane to obtain 2D vertex coordinates as observation, and add Gaussian perturbation (variance of 1 pixel) to each of the four corners as observation noise. The bounding box observation is defined by the circumscribed rectangle of the four vertices.

We triangulate each pair of 2D observations using Direct linear transformation (DLT)~\cite{mvg}, and then measure the sign position error by the Euclidean distance between the centers of the estimated and ground truth 3D corners. Figure~\ref{fig:14} shows the simulation results on 2D orientation changes caused by varying camera rotation around z-axis. In the results we can see that mapping based on bounding box detection yields larger errors as the 2D orientation change increases, while mapping based on boundary estimation shows accurate and stable estimation regardless of the 2D orientation change. This indicates that boundary estimation is crucial for robust mapping of 3D objects from 2D observations.

\section{CONCLUSIONS}
In this paper, we proposed the efficient traffic sign detection method where the locations of traffic signs are estimated together with their precise boundaries. To this end, we extended object bounding box detection problem to object pose estimation problem, and the problem is effectively modeled by CNN based on recent advances in object detection networks. The estimated pose of traffic signs is used to transform the boundary of traffic sign templates into the input image plane, providing precise boundaries with high accuracy. To achieve practical detection speed, we explored the best-performing base networks and pruned unnecessary layers of the network by considering the characteristics of traffic signs. In addition, by optimizing the resolution of network input for the best trade-off between speed and accuracy, our detector can run with frame rate of 7 FPS on low-power mobile platforms.

Future direction of our method is summarized as follows. For better accuracy, we can adopt the latest architectures of object detection such as feature pyramid network~\cite{FPN} and multi-scale training~\cite{YOLOv2}. Since the base network is a dominant factor in speed-accuracy trade-off, developing the base network specialized for traffic sign detection, instead of using the existing base network for generic object detection, is worth to study. Finally, the proposed method can be applied not only to traffic sign but also to any other planar objects having standard shapes, thus generalization of our method will be investigated.

\ifCLASSOPTIONcaptionsoff
  \newpage
\fi


\bibliographystyle{IEEEtran}
\bibliography{TSDbib}

\begin{thebibliography}{10}
\providecommand{\url}[1]{#1}
\csname url@rmstyle\endcsname
\providecommand{\newblock}{\relax}
\providecommand{\bibinfo}[2]{#2}
\providecommand\BIBentrySTDinterwordspacing{\spaceskip=0pt\relax}
\providecommand\BIBentryALTinterwordstretchfactor{4}
\providecommand\BIBentryALTinterwordspacing{\spaceskip=\fontdimen2\font plus
\BIBentryALTinterwordstretchfactor\fontdimen3\font minus
  \fontdimen4\font\relax}
\providecommand\BIBforeignlanguage[2]{{%
\expandafter\ifx\csname l@#1\endcsname\relax
\typeout{** WARNING: IEEEtran.bst: No hyphenation pattern has been}%
\typeout{** loaded for the language `#1'. Using the pattern for}%
\typeout{** the default language instead.}%
\else
\language=\csname l@#1\endcsname
\fi
#2}}

\bibitem{mapping}
O.~Dabeer, R.~Gowaikar, S.~K. Grzechnik, M.~J. Lakshman, G.~Reitmayr,
  K.~Somasundaram, R.~T. Sukhavasi, and X.~Wu, ``An end-to-end system for
  crowdsourced 3d maps for autonomous vehicles: The mapping component,'' in
  \emph{Proc. {IEEE/RSJ} International Conference on Intelligent Robots and
  Systems}, 2017.

\bibitem{segnet}
V.~Badrinarayanan, A.~Kendall, and R.~Cipolla, ``Segnet: A deep convolutional
  encoder-decoder architecture for robust semantic pixel-wise labelling,''
  \emph{ar{X}iv:1511.00561}, 2015.

\bibitem{seg_Uhrig}
J.~Uhrig, M.~Cordts, U.~Franke, and T.~Brox, ``Pixel-level encoding and depth
  layering for instance-level semantic labeling,'' \emph{ar{X}iv:1604.05096},
  2016.

\bibitem{seg_Lin}
G.~Lin, C.~Shen, A.~van~den Hengel, and I.~Reid, ``Exploring context with deep
  structured models for semantic segmentation,'' \emph{ar{X}iv:1603.03183},
  2016.

\bibitem{RCNN}
R.~Girshick, J.~Donahue, T.~Darrell, , and J.~Malik, ``Rich feature hierarchies
  for accurate object detection and semantic segmentation,'' in \emph{Proc.
  Computer Vision and Pattern Recognition}, 2014.

\bibitem{SSD}
W.~Liu, D.~Anguelov, D.~Erhan, C.~Szegedy, S.~Reed, C.-Y. Fu, and A.~C. Berg,
  ``Ssd: Single shot multibox detector,'' in \emph{Proc. European Conference on
  Computer Vision}, 2016.

\bibitem{FRCNN}
S.~Ren, K.~He, R.~Girshick, and J.~Sun, ``Faster {R-CNN}: Towards real-time
  object detection with region proposal networks,'' in \emph{Proc. Advances in
  Neural Information Processing Systems}, 2015.

\bibitem{GTSDB}
S.~Houben, J.~Stallkamp, J.~Salmen, M.~Schlipsing, and C.~Igel, ``Detection of
  traffic signs in real-world images: The {G}erman {T}raffic {S}ign {D}etection
  {B}enchmark,'' in \emph{Proc. International Joint Conference on Neural
  Networks}, 2013.

\bibitem{Yang_2016}
Y.~Yang, H.~Luo, H.~Xu, and F.~Wu, ``Towards real-time traffic sign detection
  and classification,'' \emph{{IEEE} Transactions on Intelligent Transportation
  Systems}, 2016.

\bibitem{color_prob}
Y.~Yang and F.~Wu, ``Real-time traffic sign detection via color probability
  model and integral channel features,'' \emph{Pattern Recognition}, 2014.

\bibitem{tsdr_multiple}
A.~Gudigar, C.~Shreesha, U.~Raghavendra, and U.~R. Acharya, ``Multiple
  thresholding and subspace based approach for detection and recognition of
  traffic sign,'' \emph{Multimedia Tools and Applications}, 2017.

\bibitem{Wang_2015}
D.~Wang, S.~Yue, J.~Xu, X.~Hou, , and C.-L. Liu, ``A saliency-based cascade
  method for fast traffic sign detection,'' in \emph{Proc. Intelligent Vehicles
  Symposium}, 2015.

\bibitem{tsd_occ}
C.~Liu, F.~Chang, and C.~Liu, ``Occlusion-robust traffic sign detection via
  cascaded colour cubic feature,'' \emph{{IET} Intelligent Transport Systems},
  2015.

\bibitem{MN_LBP}
C.~Liu, F.~Chang, , and Z.~Chen, ``Rapid multiclass traffic sign detection in
  high-resolution images,'' \emph{{IEEE} Transactions on Intelligent
  Transportation Systems}, 2014.

\bibitem{Wang_2013}
G.~Wang, G.~Ren, Z.~Wu, Y.~Zhao, and L.~Jiang, ``A robust, coarse-to-fine
  traffic sign detection method,'' in \emph{Proc. International Joint
  Conference on Neural Networks}, 2013.

\bibitem{ICF}
P.~Dollár, Z.~Tu, P.~Perona, and S.~Belongie, ``Integral channel features,''
  in \emph{Proc. British Machine Vision Conference}, 2009.

\bibitem{ACF}
P.~Dollar, R.~Appel, S.~Belongie, and P.~Perona, ``Fast feature pyramids for
  object detection,'' \emph{{IEEE} Transactions on Pattern Analysis and Machine
  Intelligence}, 2014.

\bibitem{TSD_US}
A.~Møgelmose, D.~Liu, and M.~M. Trivedi, ``Detection of {U.S.} traffic
  signs,'' \emph{{IEEE} Transactions on Intelligent Transportation Systems},
  2015.

\bibitem{TSDTR_ieee_ITS}
Y.~Yuan, Z.~Xiong, and Q.~Wang, ``An incremental framework for video-based
  traffic sign detection, tracking and recognition,'' \emph{{IEEE} Transactions
  on Intelligent Transportation Systems}, 2016.

\bibitem{Alex}
A.~Krizhevsky, I.~Sutskever, and G.~E. Hinton, ``Imagenet classification with
  deep convolutional neural networks,'' in \emph{Proc. Advances in Neural
  Information Processing Systems}, 2012.

\bibitem{TSC_idsia}
D.~Ciresan, U.~Meier, J.~Masci, and J.~Schmidhuber, ``Multi-column deep neural
  network for traffic sign classification,'' \emph{Neural Networks}, 2012.

\bibitem{Wu_2013}
Y.~Wu, Y.~Liu, J.~Li, H.~Liu, and X.~Hu, ``Traffic sign detection based on
  convolutional neural networks,'' in \emph{Proc. International Joint
  Conference on Neural Networks}, 2013.

\bibitem{Qian_2015}
R.~Qian, B.~Zhang, Y.~Yue, Z.~Wang, and F.~Coenen, ``Robust chinese traffic
  sign detection and recognition with deep convolutional neural network,'' in
  \emph{Proc. International Conference on Natural Computation}, 2015.

\bibitem{tsdr_review}
A.~Gudigar, C.~Shreesha, and U.~Raghavendra, ``A review on automatic detection
  and recognition of traffic sign,'' \emph{Multimedia Tools and Applications},
  2016.

\bibitem{tsrd_review2}
S.~B. Wali, M.~A. Hannan, A.~Hussain, and S.~A. Samad, ``Comparative survey on
  traffic sign detection and recognition: a review,'' \emph{Przegld
  Elektrotechniczny}, 2015.

\bibitem{tsrd_review3}
P.~Saxena, N.~Gupta, S.~Y. Laskar, and P.~P. Borah, ``A study on automatic
  detection and recognition techniques for road signs,'' \emph{International
  Journal of Computational Engineering Research}, 2015.

\bibitem{yolo}
J.~Redmon, S.~Divvala, R.~Girshick, , and A.~Farhadi, ``You only look once:
  Unified, real-time object detection,'' in \emph{Proc. Computer Vision and
  Pattern Recognition}, 2016.

\bibitem{rfcn}
Y.~Li, K.~He, J.~Sun, \emph{et~al.}, ``{R-FCN}: Object detection via
  region-based fully convolutional networks,'' in \emph{Proc. Advances in
  Neural Information Processing Systems}, 2016.

\bibitem{speed}
J.~Huang, V.~Rathod, C.~Sun, M.~Zhu, A.~Korattikara, A.~Fathi, I.~Fischer,
  Z.~Wojna, Y.~Song, S.~Guadarrama, \emph{et~al.}, ``Speed/accuracy trade-offs
  for modern convolutional object detectors,'' \emph{arXiv preprint
  arXiv:1611.10012}, 2016.

\bibitem{TSD_wild}
Z.~Zhu, H.-Z. Huang, Z.-P. Tan, K.~Xu, and S.-M. Hu, ``Traffic-sign detection
  and classification in the wild,'' in \emph{Proc. Computer Vision and Pattern
  Recognition}, 2016.

\bibitem{overfeat}
P.~Sermanet, D.~Eigen, X.~Zhang, M.~Mathieu, R.~Fergus, and Y.~LeCun,
  ``Overfeat: {I}ntegrated recognition, localization and detection using
  convolutional networks,'' in \emph{Proc. International Conference on Learning
  Representations}, 2014.

\bibitem{vgg}
K.~Simonyan and A.~Zisserman, ``Very deep convolutional networks for
  large-scale image recognition,'' in \emph{Proc. International Conference on
  Learning Representations}, 2015.

\bibitem{inception}
C.~Szegedy, W.~Liu, Y.~Jia, P.~Sermanet, S.~Reed, D.~Anguelov, D.~Erhan,
  V.~Vanhoucke, and A.~Rabinovich, ``Going deeper with convolutions,'' in
  \emph{Proc. Computer Vision and Pattern Recognition}, 2015.

\bibitem{Resnet}
K.~He, X.~Zhang, S.~Ren, and J.~Sun, ``Deep residual learning for image
  recognition,'' in \emph{Proc. Computer Vision and Pattern Recognition}, 2016.

\bibitem{decaf}
J.~Donahue, Y.~Jia, O.~Vinyals, J.~Hoffman, N.~Zhang, E.~Tzeng, and T.~Darrell,
  ``{DeCAF}: A deep convolutional activation feature for generic visual
  recognition.'' in \emph{Proc. International Conference on Machine Learning},
  2014.

\bibitem{base_comp}
J.~Huang, V.~Rathod, C.~Sun, M.~Zhu, A.~Korattikara, A.~Fathi, I.~Fischer,
  Z.~Wojna, Y.~Song, S.~Guadarrama, and K.~Murphy, ``Speed/accuracy trade-offs
  for modern convolutional object detectors,'' \emph{ar{X}iv:1611.10012}, 2016.

\bibitem{MUTCD}
{California State Transportation Agency, Department of Transportation},
  \emph{California Manual on Uniform Traffic Control Devices}.\hskip 1em plus
  0.5em minus 0.4em\relax CA, USA, 2014.

\bibitem{imagenet}
J.~Deng, W.~Dong, R.~Socher, L.-J. Li, K.~Li, and L.~Fei-Fei, ``{ImageNet: A
  Large-Scale Hierarchical Image Database},'' in \emph{Proc. Computer Vision
  and Pattern Recognition}, 2009.

\bibitem{Xavier}
X.~Glorot and Y.~Bengio, ``Understanding the difficulty of training deep
  feedforward neural networks,'' in \emph{Proc. International Conference on
  Artificial Intelligence and Statistics}, 2010.

\bibitem{fastrcnn}
R.~Girshick, ``Fast {R-CNN},'' in \emph{Proc. {IEEE} International Conference
  on Computer Vision}, 2015.

\bibitem{hardnegative}
A.~Shrivastava, A.~Gupta, and R.~Girshick, ``Training region-based object
  detectors with online hard example mining,'' in \emph{Proc. Computer Vision
  and Pattern Recognition}, 2016.

\bibitem{LISA}
A.~Møgelmose, M.~M. Trivedi, and T.~B. Moeslund, ``Vision based traffic sign
  detection and analysis for intelligent driver assistance systems:
  Perspectives and survey,'' \emph{{IEEE} Transactions on Intelligent
  Transportation Systems}, 2012.

\bibitem{caffe}
Y.~Jia, E.~Shelhamer, J.~Donahue, S.~Karayev, J.~Long, R.~Girshick,
  S.~Guadarrama, and T.~Darrell, ``{Caffe}: Convolutional architecture for fast
  feature embedding,'' \emph{ar{X}iv:1408.5093}, 2014.

\bibitem{visics}
M.~Mathias, R.~Timofte, R.~Benenson, and L.~V. Gool, ``Traffic sign recognition
  - how far are we from the solution?'' in \emph{Proc. {IEEE} International
  Joint Conference on Neural Networks}, 2013.

\bibitem{LITS1}
M.~Liang, M.~Yuan, X.~Hu, J.~Li, and H.~Liu, ``Traffic sign detection by roi
  extraction and histogram features-based recognition,'' in \emph{Proc. {IEEE}
  International Joint Conference on Neural Networks}, 2013.

\bibitem{BolognaCVLab}
S.~Salti, A.~Petrelli, F.~Tombari, N.~Fioraio, and L.~D. Stefano, ``A traffic
  sign detection pipeline based on interest region extraction,'' in \emph{Proc.
  {IEEE} International Joint Conference on Neural Networks}, 2013.

\bibitem{InceptionV2}
S.~Ioffe and C.~Szegedy, ``Batch normalization: Accelerating deep network
  training by reducing internal covariate shift,'' in \emph{Proc. International
  Conference on Machine Learning}, 2015.

\bibitem{mvg}
R.~I. Hartley and A.~Zisserman, \emph{Multiple View Geometry in Computer
  Vision}, 2nd~ed.\hskip 1em plus 0.5em minus 0.4em\relax Cambridge University
  Press, 2004.

\bibitem{FPN}
T.-Y. Lin, P.~Dollar, R.~Girshick, K.~He, B.~Hariharan, and S.~Belongie,
  ``Feature pyramid networks for object detection,''
  \emph{ar{X}iv:1612.03144v1}, 2016.

\bibitem{YOLOv2}
J.~Redmon and A.~Farhadi, ``{YOLO9000}: Better, faster, stronger,'' in
  \emph{Proc. Computer Vision and Pattern Recognition}, 2017.

\end{thebibliography}

%

\begin{IEEEbiographynophoto}{Hee Seok Lee}
received the BS and PhD degrees in electrical engineering from Seoul National University, Seoul, Korea, in 2006 and 2013, respectively. During the PhD study, he was a research assistant for the Computer Vision Laboratory. He is currently with Qualcomm Korea research center, Seoul, Korea. His research interests are in vision-based SLAM and 3D reconstruction, visual tracking, and object detection.
\end{IEEEbiographynophoto}

\begin{IEEEbiographynophoto}{Kang Kim}
received the BS degree in computer science from Yonsei University, Seoul, Korea in 2007 and obtained the MS degree in computer science from KAIST, Daejeon, Korea in 2009.  He is currently with Qualcomm Korea research center, Seoul, Korea. His research interests include object detection, semantic segmentation, and scene text recognition using deep learning.
\end{IEEEbiographynophoto}

\newpage




\end{document}